\documentclass{llncs}

\usepackage{makeidx}  
\usepackage{times}
\usepackage{graphicx} 
\usepackage{subfigure} 

\usepackage{natbib}

\usepackage{algorithm}
\usepackage{algorithmic}

\usepackage{xcolor,colortbl}

\usepackage{hyperref}

\renewcommand{\bibsection}{\section*{\refname}\small\renewcommand\bibnumfmt[1]{##1.}}

\usepackage{amsmath,amssymb,amsfonts,mathtools}
\usepackage{enumitem}
\usepackage{multirow}
\usepackage{array}
\usepackage{bm}
\usepackage{color}

\DeclareMathOperator*{\minimize}{minimize}
\DeclareMathOperator*{\st}{s.t.}

\def\AlgoSize{}

\def\reals{\mathbb{R}}
\newcommand{\R}{\mathbb{R}}

\def\w{\bm w}
\def\W{\bm W}
\def\V{\bm V}

\def\U{\bm U}
\def\I{\bm I}
\def\diag{\text{diag}}
\def\G{\mathcal{G}}

\def\L{\mathcal{L}}

\def\F{\text{F}}

\newlength{\widebarargwidth}
\newlength{\widebarargheight}
\newlength{\widebarargdepth}
\DeclareRobustCommand{\widebar}[1]{%
  \settowidth{\widebarargwidth}{\ensuremath{#1}}%
  \settoheight{\widebarargheight}{\ensuremath{#1}}%
  \settodepth{\widebarargdepth}{\ensuremath{#1}}%
  \addtolength{\widebarargwidth}{-0.3\widebarargheight}%
  \addtolength{\widebarargwidth}{-0.3\widebarargdepth}%
  \makebox[0pt][l]{\hspace{0.32\widebarargheight}%
    \hspace{0.3\widebarargdepth}%
    \addtolength{\widebarargheight}{0.3ex}%
    \rule[\widebarargheight]{1\widebarargwidth}{0.1ex}}%
  {#1}}

\usepackage{booktabs}

\begin{document} 

\pagestyle{headings}

\mainmatter              

\title{Learning task structure via sparsity grouped multitask learning}

\titlerunning{Sparsity grouped multitask learning}

\author{Meghana Kshirsagar\inst{1} \and Eunho Yang\inst{2} \and Aur\'{e}lie C. Lozano\inst{3}}

\institute{\footnote{\scriptsize{This work was done while MK and EY were at IBM T.J. Watson research}}Memorial Sloan Kettering Cancer Center,\\
       1275 York Ave., New York, NY\\
       \email{kshirsam@mskcc.org}
 \and
       School of Computing,\\
       Korea Advanced Inst. of Science and Tech., Daejeon, South Korea.\\
       \email{eunhoy@kaist.ac.kr}
\and
       IBM T. J. Watson research\\
       Yorktown Heights, New York, NY\\
       \email{aclozano@us.ibm.com}
}



\maketitle              

\begin{abstract}
Sparse mapping has been a key methodology in many high-dimensional scientific problems. When multiple tasks share the set of relevant features, learning them jointly in a group drastically improves the quality of relevant feature selection. However, in practice this technique is used limitedly since such grouping information is usually hidden. In this paper, our goal is to recover the group structure on the sparsity patterns and leverage that information in the sparse learning. Toward this, we formulate a joint optimization problem in the task parameter and the group membership, by constructing an appropriate regularizer to encourage sparse learning as well as correct recovery of task groups. We further demonstrate that our proposed method recovers groups and the sparsity patterns in the task parameters accurately by extensive experiments. 
\end{abstract}

\section{Introduction}

Humans acquire knowledge and skills by categorizing the various problems/tasks encountered, recognizing how the tasks are related to each other and taking advantage of this organization when learning a new task. Statistical machine learning methods also benefit from exploiting such similarities in learning related problems. 
Multitask learning (MTL) \citep{caruana1997} is a paradigm of machine learning, encompassing learning algorithms that can share information among related tasks and help to perform those tasks together more efficiently than in isolation.
These algorithms exploit task relatedness by various mechanisms. Some works enforce that parameters of various tasks are close to each other in some geometric sense \citep{pontil2004,maurer2006}. Several works leverage the existence of a shared low dimensional subspace \citep{argyriou2008,liu2009,jalali2010,chen2012} or manifold \citep{agarwal2010} that contains the task parameters. Some bayesian MTL methods assume the same prior on parameters of related tasks \citep{yu2005,daume2009}, while neural networks based methods share some hidden units \citep{baxter2000}. 

A key drawback of most MTL methods is that they assume that all tasks are equally related. Intuitively, learning unrelated tasks jointly may result in poor predictive models; i.e tasks should be coupled based on their relatedness. While the coupling of task parameters can sometimes be controlled via hyper-parameters, this is infeasible when learning several hundreds of tasks. Often, knowing the \textit{task relationships} themselves is of interest to the application at hand. While these relationships might sometimes be derived from domain specific intuition \citep{kim2010,widmer,rao2013sparse}, they are either not always known apriori or are pre-defined based on the knowledge of $P(X)$ rather than $P(Y|X)$. We aim to automatically learn these task relationships, while simultaneously learning individual task parameters. This idea of jointly learning task groups and parameters has been explored in prior works. For instance \citet{argyriou2008} learn a set of kernels, one per group of tasks and \citet{jacob2009} cluster tasks based on similarity of task parameters. Others~ \citep{zhang2010,gong2012} try to identify ``outlier'' tasks. \citet{kumar2012,kang2011} assume that task parameters within a group lie in a shared low dimensional subspace. \citet{yizhang2010} use a matrix-normal regularization to capture task covariance and feature covariance between tasks and enforce sparsity on these covariance parameters and \citet{fei2013} use a similar objective with a structured regularizer. Their approach is however, not suited for high dimensional settings and they do not enforce any sparsity constraints on the task parameters matrix $W$. A Bayesian approach is proposed in \citet{passos2012}, where parameters are assumed to come from a nonparametric mixture of nonparametric factor analyzers.

Here, we explore the notion of \textit{shared sparsity} as the structure connecting a group of related tasks. More concretely, we assume that tasks in a group all have similar relevant features or analogously, the same zeros in their parameter vectors. Sparsity inducing norms such as the $\ell_1$ norm capture the principle of parsimony, which is important to many real-world applications, and have enabled efficient learning in settings with high dimensional feature spaces and few examples, via algorithms like the Lasso \citep{tibshirani1996}. When confronted by several tasks where sparsity is required, one modeling choice is for each task to have its' own sparse parameter vector. At the other extreme is the possibility of enforcing shared sparsity on all tasks via a structured sparsity inducing norm such as $\ell_1/\ell_2$ on the task parameter matrix: $\|\W\|_{1,2}$ \citep{bach2011}. \footnote{Note: this cross-task structured sparsity is different from the Group Lasso \citep{yuan2006}, which groups covariates within a task (${\displaystyle \min_{w \in \mathbb{R}^d}} \sum_g \|w_g\|$, where $w_g$ is a group of parameters)}. We choose to enforce sparsity at a group level by penalizing $\|\W_g\|_{1,2}$, where $\|\W_g\|$ is the parameter matrix for all tasks in group $g$, while learning group memberships of tasks. 

To see why this structure is interesting and relevant, consider the problem of transcription factor (TF) binding prediction. TFs are proteins that bind to the DNA to regulate expression of nearby genes. The binding specificity of a TF to an arbitrary location on the DNA depends on the pattern/sequence of nucleic acids (A/C/G/T) at that location. These sequence preferences of TFs have some similarities among related TFs. Consider the task of predicting TF binding, given segments of DNA sequence (these are the examples), on which we have derived features such as $n$-grams (called $k$-mers) \footnote{e.g.: GTAATTNC is an 8-mer (`N' represents a wild card)}. The feature space is very high dimensional and a small set of features typically capture the binding pattern for a single TF. Given several tasks, each representing one TF, one can see that the structure of the ideal parameter matrix is likely to be group sparse, where TFs in a group have similar binding patterns (i.e similar important features but with different weights). The applicability of task-group based sparsity is not limited to isolated applications, but desirable in problems involving billions of features, as is the case with web-scale information retrieval and in settings with few samples such as genome wide association studies involving millions of genetic markers over a few hundred patients, where only a few markers are relevant.

The main contributions of this work are: 
\begin{itemize}
\item We present a new approach towards learning task group structure in a multitask learning setting that simultaneously learns both the task parameters $W$ and a clustering over the tasks. 
\item We define a regularizer that divides the set of tasks into groups such that all tasks within a group share the same sparsity structure. Though the ideal regularizer is discrete, we propose a relaxed version and we carefully make many choices that lead to a feasible alternating minimization based optimization strategy. We find that several alternate formulations result in substantially worse solutions.
\item We evaluate our method through experiments on synthetic datasets and two interesting real-world problem settings. The first is a regression problem: QSAR, quantitative structure activity relationship prediction (see \citep{ma2015} for an overview) and the second is a classification problem important in the area of regulatory genomics: transcription factor binding prediction (described above). On synthetic data with known group structure, our method recovers the correct structure. On real data, we perform better than prior MTL group learning baselines.
\end{itemize}

\subsection{Relation to prior work}
Our work is most closely related to \citet{kang2011}, who assume that each group of tasks shares a latent subspace. They find groups so that $\|\W_g\|_*$ for each group $g$ is small, thereby enforcing sparsity in a \textit{transformed} feature space. Another approach, GO-MTL \citep{kumar2012} is based on the same idea, with the exception that the latent subspace is shared among all tasks, and a low-rank decomposition of the parameter matrix $\W = {\bm L}{\bm S}$ is learned. Subsequently, the coefficients matrix $S$ is clustered to obtain a grouping of tasks. Note that, learning group memberships is not the goal of their approach, but rather a post-processing step upon learning their model parameters.

To understand the distinction from prior work, consider the weight matrix $\W^*$ in Figure \ref{FigToyComp}(a), which represents the true group sparsity structure that we wish to learn. While each task group has a low-rank structure (since $s$ of the $d$ features are non-zero, the rank of any $\W_g$ is bounded by $s$), it has an additional property that ($d-s$) features are zero or irrelevant for all tasks in this group. Our method is able to exploit this additional information to learn the correct sparsity pattern in the groups, while that of \citet{kang2011} is unable to do so, as illustrated on this synthetic dataset in Figure \ref{FigToyKang} (details of this dataset are in Sec \ref{sec:synthdata}). Though \citet{kang2011} recovers some of the block diagonal structure of $W$, there are many non-zero features which lead to an incorrect group structure. We present a further discussion on how our method is sample efficient as compared to \citet{kang2011} for this structure of $\W$ in Sec \ref{sec:complexity}.

We next present the setup and notation, and lead to our approach by starting with a straight-forward combinatorial objective and then make changes to it in multiple steps (Sec 2-4). At each step we explain the behaviour of the function to motivate the particular choices we made; present a high-level analysis of sample complexity for our method and competing methods. Finally we show experiments (Sec 5) on four datasets.

\section{Setup and Motivation}
 
We consider the standard setting of multi-task learning in particular where tasks in the same group share the sparsity patterns on parameters. Let $\{T_1, \hdots, T_m\}$  be the set of $m$ tasks with training data $\mathcal{D}_t$ ($t$ = $1 \hdots m$). Let the parameter vectors corresponding to each of the $m$ tasks be $\w^{(1)}, \w^{(2)}, \hdots, \w^{(m)} \in \reals^{d}$, $d$ is the number of covariates/features. Let $\L(\cdot)$ be the loss function which, given $\mathcal{D}_t$ and $\w^{(t)}$ measures deviation of the predictions from the response. Our goal is to learn the task parameters where i) each $\w^{(t)}$ is assumed to be sparse so that the response of the task can be succinctly explained by a small set of features, and moreover ii) there is a partition $\G^* := \{G_1, G_2, \hdots, G_{N}\}$ over tasks such that all tasks in the same group $G_i$ have the same sparsity patterns. Here $N$ is the total number of groups learned. If we learn every task independently, we solve $m$ independent optimization problems:
\begin{align*}
	\minimize_{\w^{(t)} \in \reals^d} \L(\w^{(t)}; \mathcal{D}_t) + \lambda \| \w^{(t)} \|_1
\end{align*} 
where $\| \w^{(t)} \|_1$ encourages sparse estimation with regularization parameter $\lambda$. However, if $\G^*$ is given, jointly estimating all parameters together using a group regularizer (such as $\ell_1/\ell_2$ norm), is known to be more effective. This approach requires fewer samples to recover the sparsity patterns by sharing information across tasks in a group:   
\begin{align} \label{EqnGroupLasso}
	\minimize_{\w^{(1)}, \hdots, \w^{(m)}} \, \sum_{t=1}^{m} \L(\w^{(t)}; \mathcal{D}_t)  + \sum_{g \in \G^*} \lambda_g \big\| \W_g \big\|_{1,2}
\end{align}
where $\W_g \in \R^{d \times |g|}$, where $|g|$ is the number of tasks in the group $g$ and $\|\cdot\|_{1,2}$ is the sum of $\ell_2$ norms computed over row vectors. Say $t_1, t_2 \hdots t_k$ belong to group $g$, then $\big\| \W_g \big\|_{1,2} := \sum_{j=1}^d \sqrt{(w^{(t_1)}_j)^2 + (w^{(t_2)}_j)^2 + \hdots + (w^{(t_k)}_j)^2}$. Here $w^{(t)}_j$ is the $j$-th entry of vector $\w^{(t)}$. Note that here we use $\ell_2$ norm for grouping, but any $\ell_\alpha$ norm $\alpha \geq 2$  is known to be effective. 

\begin{figure}[t]
	\centering
	\begin{tabular}{cc}
		\hspace{-.45cm}\includegraphics[width=0.28\textwidth]{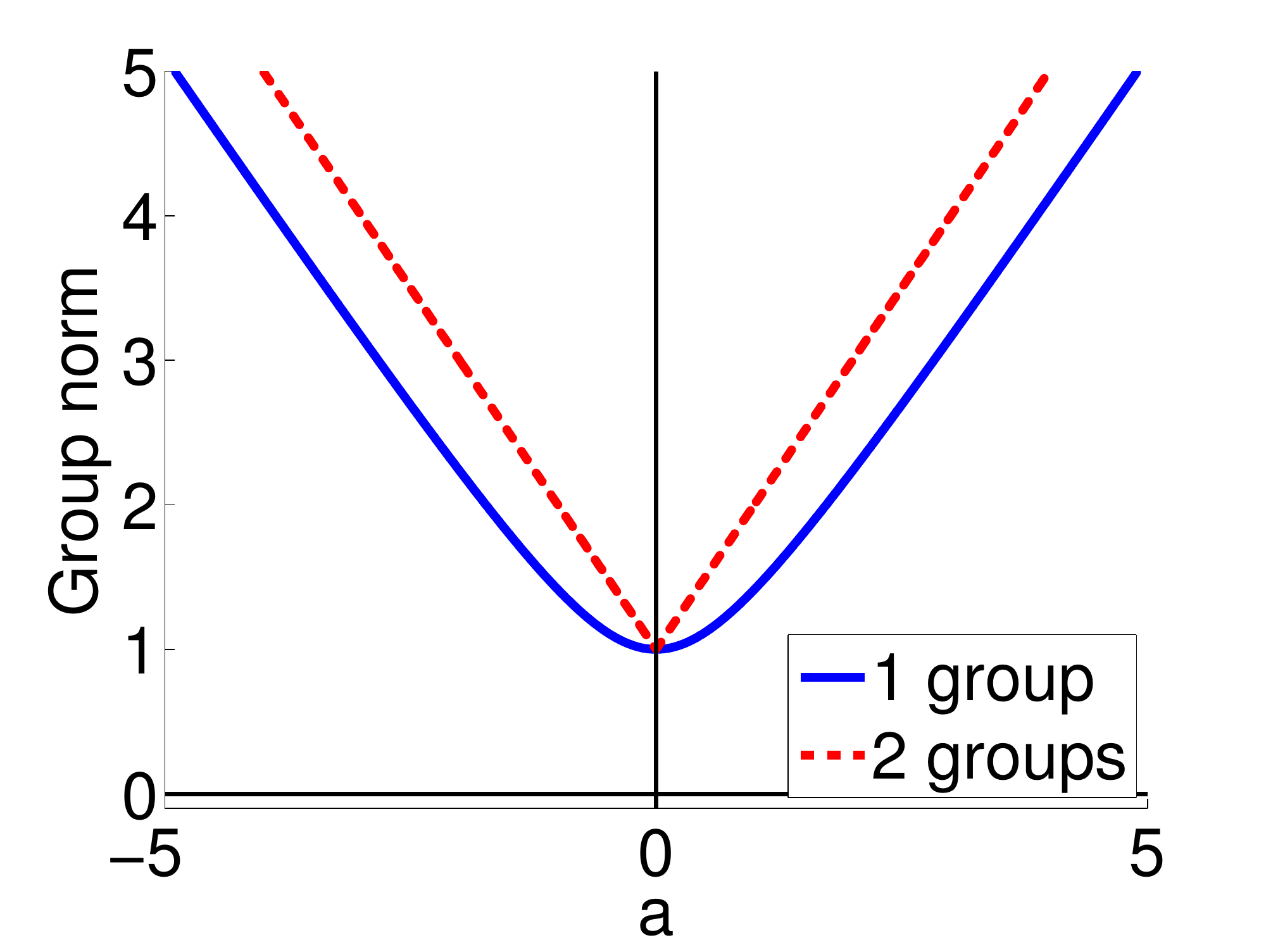} & \hspace{-.4cm}\includegraphics[width=0.28\textwidth]{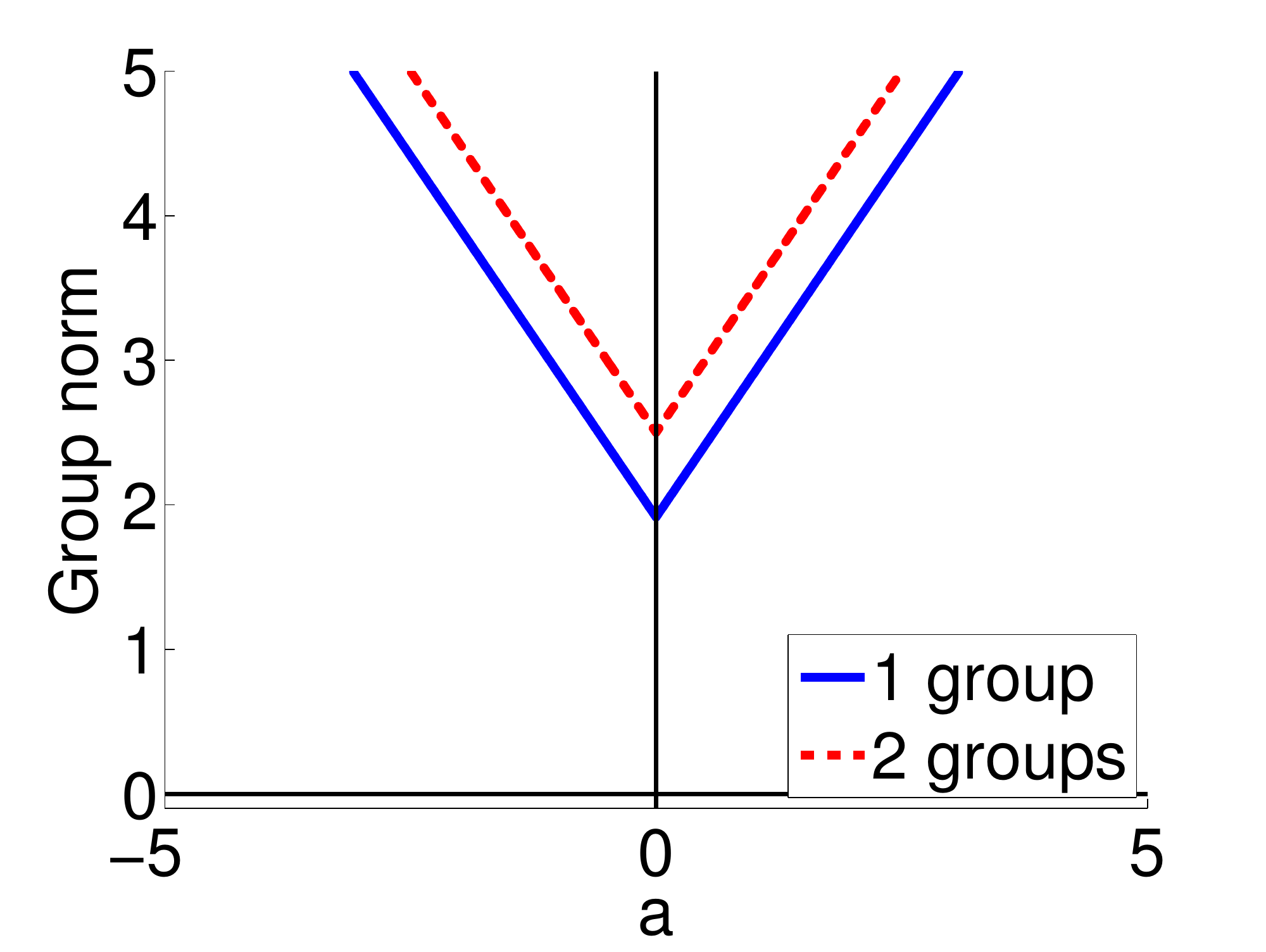}
	\end{tabular}
	\caption{Toy examples with two fixed parameter vectors: (Left) $\w^{(1)} = (1,0,0)^\top$, $\w^{(2)} = (a,0,0)^\top$, and (Right) $\w^{(1)} = (1/2,1,0)^\top$, $\w^{(2)} = (0,1,a)^\top$, where we only vary one coordinate $a$ fixing all others to visualize the norm values in 2-d space. Functions show the group norms, $\sum_g \|\W\U_g\|_{1,2}$ in \eqref{EqnBaseline} where two tasks belong to a single group (solid) or to separate groups (dotted). In both cases, this group regularizer favors the case with a single group.}
	\label{FigExpL1}
\end{figure}

We introduce a membership parameter $u_{g,t}$: $u_{g,t} = 1$ if task $T_t$ is in a group $g$ and $0$ otherwise. Since we are only allowing a hard membership without overlapping (though this assumption can be relaxed in future work), we should have exactly one active membership parameter for each task: $u_{g,t} =1$ for some $g \in \G$ and $u_{g',t} = 0$ for all other $g' \in \G\setminus \{g\}$. For notational simplicity, we represent the group membership parameters for a group $g$ in the form of a matrix $\U_g$. This is a diagonal matrix where $\U_g := \diag (u_{g,1}, u_{g,2},\hdots, u_{g,m} ) \in \{0,1\}^{m \times m}$. In other words, $ [\U_g]_{ii} = u_{g,i} = 1$ if task $T_i$ is in group $g$ and $0$ otherwise. Now, incorporating $\U$ in \eqref{EqnGroupLasso}, we can derive the optimization problem for learning the task parameters $\{\w^{(t)}\}_{t=1,\hdots,m}$ and $\U$ simultaneously as follows:
\begin{align}\label{EqnBaseline}
	\minimize_{\W,\U} & \, \sum_{t=1}^{m} \L(\w^{(t)}; \mathcal{D}_t)  + \sum_{g \in \G} \lambda_g \big\| \W \U_g \big\|_{1,2}  \nonumber\\
	\st & \, \sum_{g \in \G} \U_g = \I^{m\times m} \ , \quad [\U_g]_{ii} \in \{0,1\} \, .
\end{align}
where $\W \in \reals^{d\times m} := [\w^{(1)}, \w^{(2)}, \hdots, \w^{(m)}]$. Here $\I^{m\times m}$ is the $m\times m$ identity matrix. After solving this problem, $\U$ encodes which group the task $T_t$ belongs to. It turns out that this simple extension in \eqref{EqnBaseline} fails to correctly infer the group structure as it is biased towards finding a smaller number of groups. 
Figure \ref{FigExpL1} shows a toy example illustrating this. The following proposition generalizes this issue.

\begin{proposition}\label{PropSumL1}
	Consider the problem of minimizing \eqref{EqnBaseline} with respect to $\U$ for a fixed $\widehat{\W}$. The assignment such that $\widehat{\U}_g =\I^{m\times m}$ for some $g \in \G$ and $\widehat{\U}_{g'} = {\bm 0}^{m\times m}$ for all other $g' \in \G\setminus \{g\}$, is a minimizer of \eqref{EqnBaseline}.
\end{proposition}
\noindent\textbf{Proof:} Please refer to the appendix.


\section{Learning Groups on Sparsity Patterns}

In the previous section, we observed that the standard group norm is beneficial when the group structure $\G^*$ is known but not suitable for inferring it. This is mainly because it is basically aggregating groups via the $\ell_1$ norm; let ${\bm v} \in \reals^{N}$ be a vector of $(\|\W\U_1\|_{1,2}, \|\W\U_2\|_{1,2}, \hdots, \|\W\U_{N}\|_{1,2})^{\top}$, then the regularizer of \eqref{EqnBaseline} can be understood as $\|{\bm v}\|_1$. By the basic property of $\ell_1$ norm, $\bm v$ tends to be a sparse vector, making $\U$ have a small number of active groups (we say some group $g$ is active if there exists a task $T_t$ such that $u_{g,t} = 1$.)

\begin{figure}[t]
	\centering
	\begin{tabular}{cc}
		\hspace{-.45cm}\includegraphics[width=0.28\textwidth]{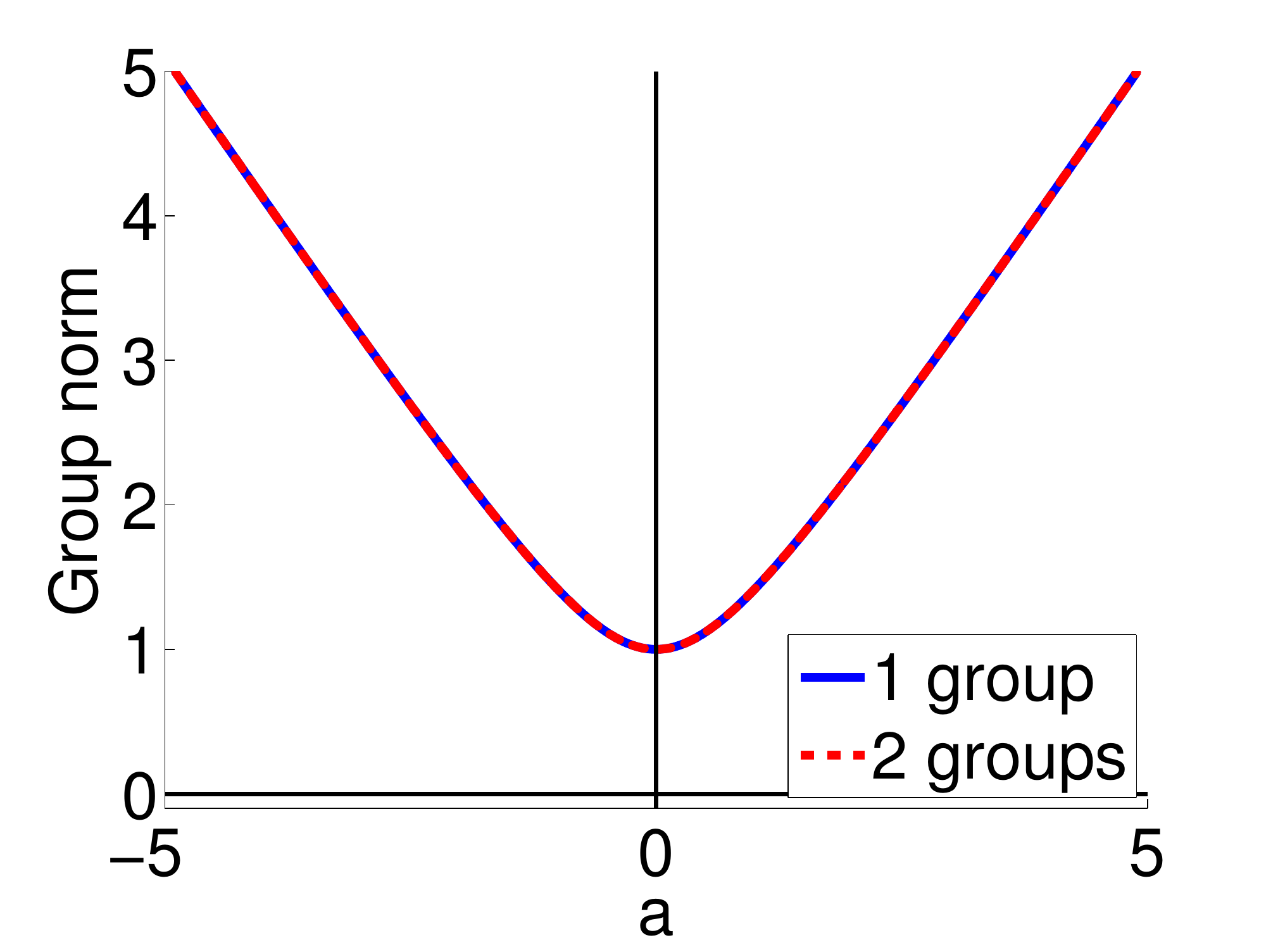} & \hspace{-.4cm}\includegraphics[width=0.28\textwidth]{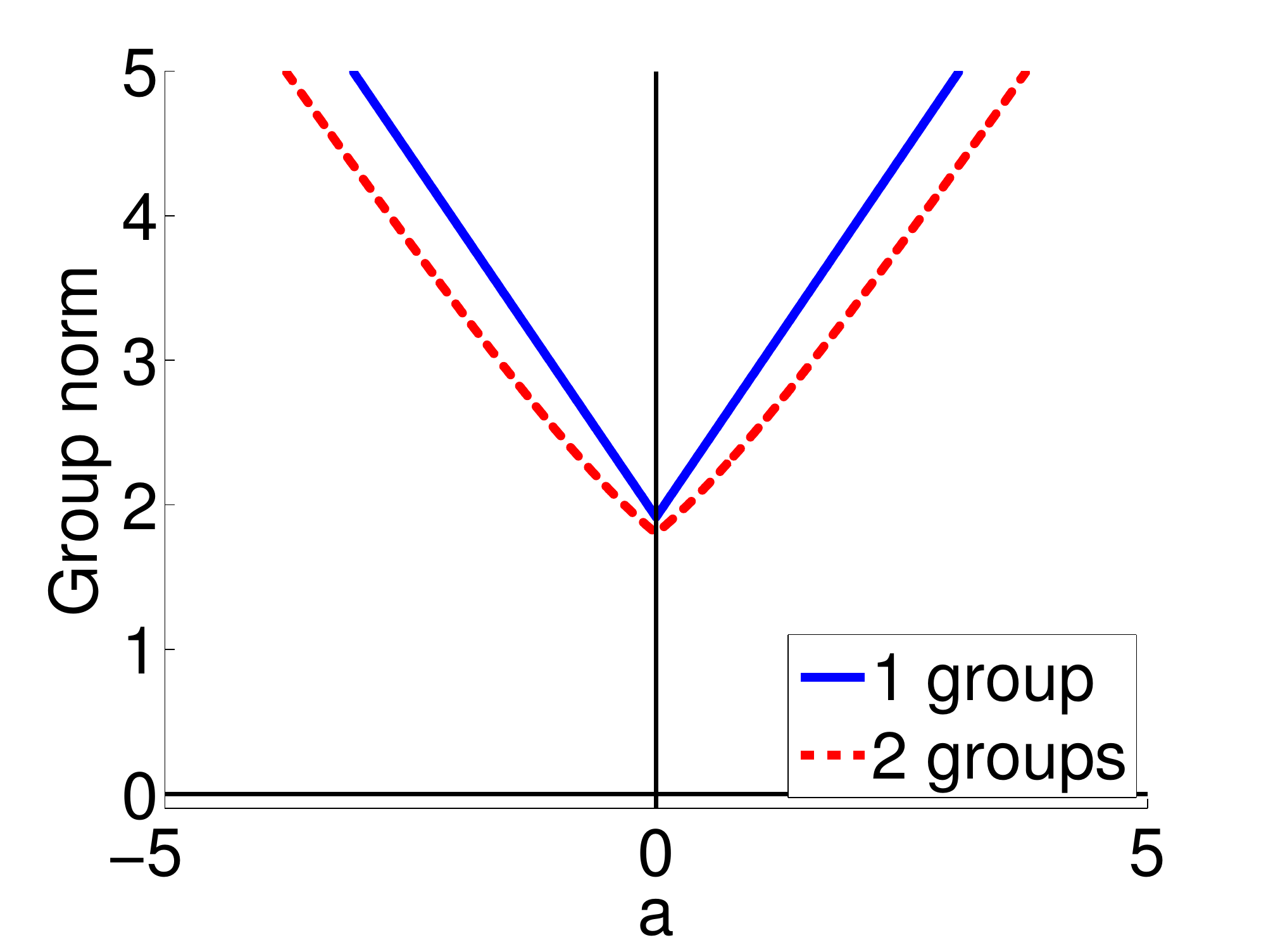}
	\end{tabular}
	\caption{For the two toy examples from Figure \ref{FigExpL1}, we show the behaviour of $\big(\sum_g (\|\W\U_g\|_{1,2})^2\big)^{0.5}$ (the group regularizer in \eqref{EqnGrpLearn1} with $\alpha=0.5$). See the caption of Figure \ref{FigExpL1} for the choice of $\W = [\w^{(1)}, \w^{(2)}]$. In the example on the right, the regularizer now favors putting the tasks in separate groups.}
	\label{FigExpL2}
\end{figure}

Based on this finding, we propose to use the $\ell_\alpha$ norm ($\alpha \geq 2$) for summing up the regularizers from different groups, so that the final regularizer as a whole forces most of $\|\W\U_g\|_{1,2}$ to be non-zeros:
\begin{align}\label{EqnGrpLearn1}
	\minimize_{\W,\U} & \, \sum_{t=1}^{m}  \L(\w^{(t)}; \mathcal{D}_t)  + \sum_{g \in \G} \lambda_g \Big(  \big\| \W \U_g \big\|_{1,2} \Big)^\alpha \nonumber\\
	\st & \, \sum_{g \in \G} \U_g = \I^{m\times m} \ , \quad [\U_g]_{ik} \in \{0,1\} \, .
\end{align}
Note that strictly speaking, $\|{\bm v}\|_\alpha$ is defined as $(\sum_{i=1}^{N} |{ v}_i|^{\alpha})^{1/\alpha}$, but we ignore the relative effect of $1/\alpha$ in the exponent. One might want to get this exponent back especially when $\alpha$ is large. $\ell_\alpha$ norms give rise to exactly the opposite effects in distributions of $u_{g,t}$, as shown in Figure \ref{FigExpL2} and Proposition \ref{PropSumGN}. 

\begin{proposition}\label{PropSumGN}
	Consider a minimizer $\widehat{\U}$ of \eqref{EqnGrpLearn1}, for any fixed $\widehat{\W}$. Suppose that there exist two tasks in a single group such that $\widehat{w}^{(s)}_i \widehat{w}^{(t)}_j \neq \widehat{w}^{(s)}_j \widehat{w}^{(t)}_i$. Then there is no empty group $g$ such that $\widehat{\U}_g ={\bm 0}^{m\times m}$.
\end{proposition}
\noindent\textbf{Proof:} Please refer to the appendix.

\begin{figure}[t]
	\centering
	\begin{tabular}{ccc}
		\hspace{-.4cm}\includegraphics[width=0.15\textwidth, clip=true,trim=3.5cm 0cm 3.5cm 1.5cm]{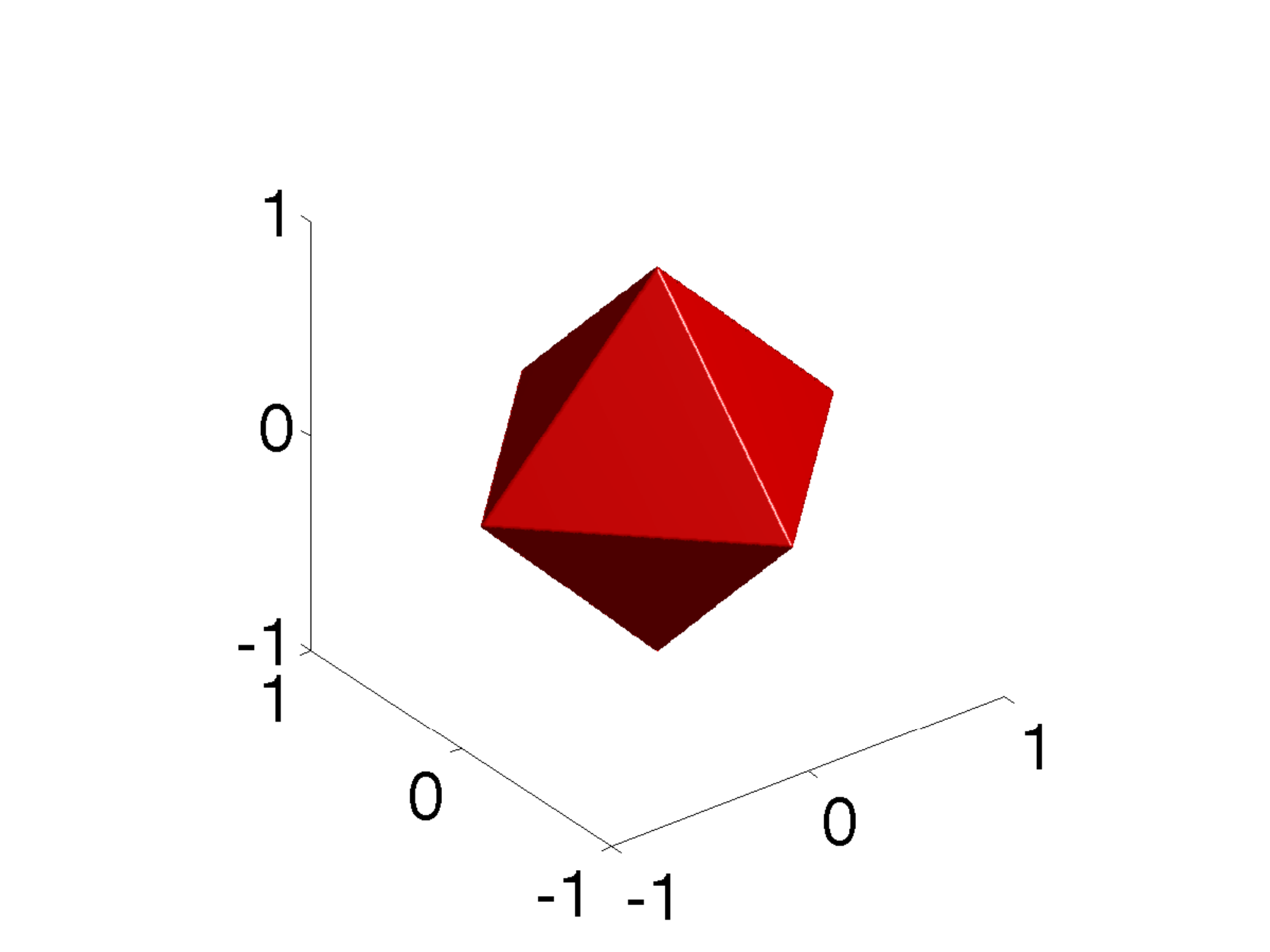} & \hspace{-.1cm}\includegraphics[width=0.15\textwidth, clip=true,trim=3.5cm 0cm 3.5cm 1.5cm]{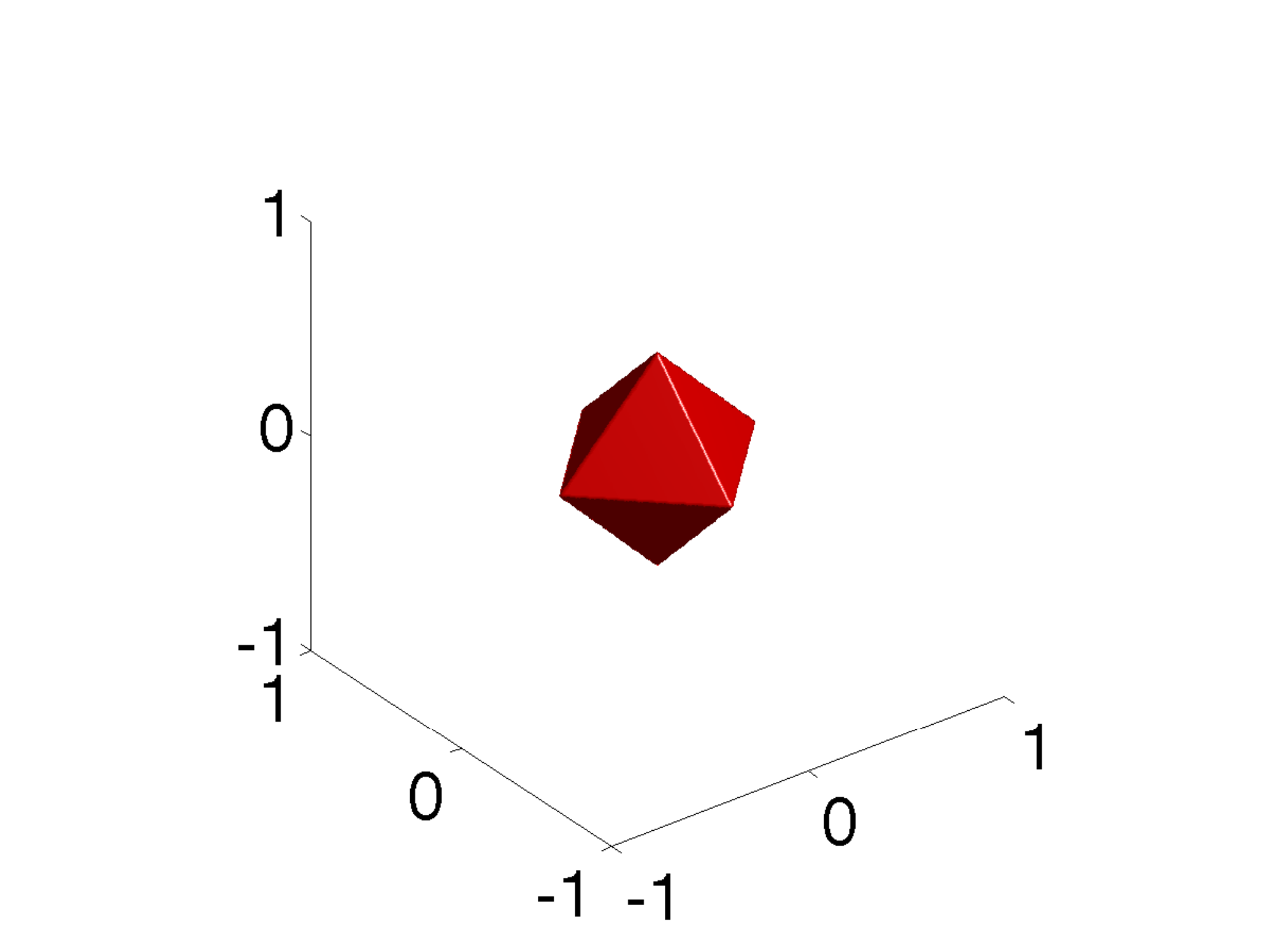} & \hspace{-.1cm}\includegraphics[width=0.15\textwidth, clip=true,trim=3.5cm 0cm 3.5cm 1.5cm]{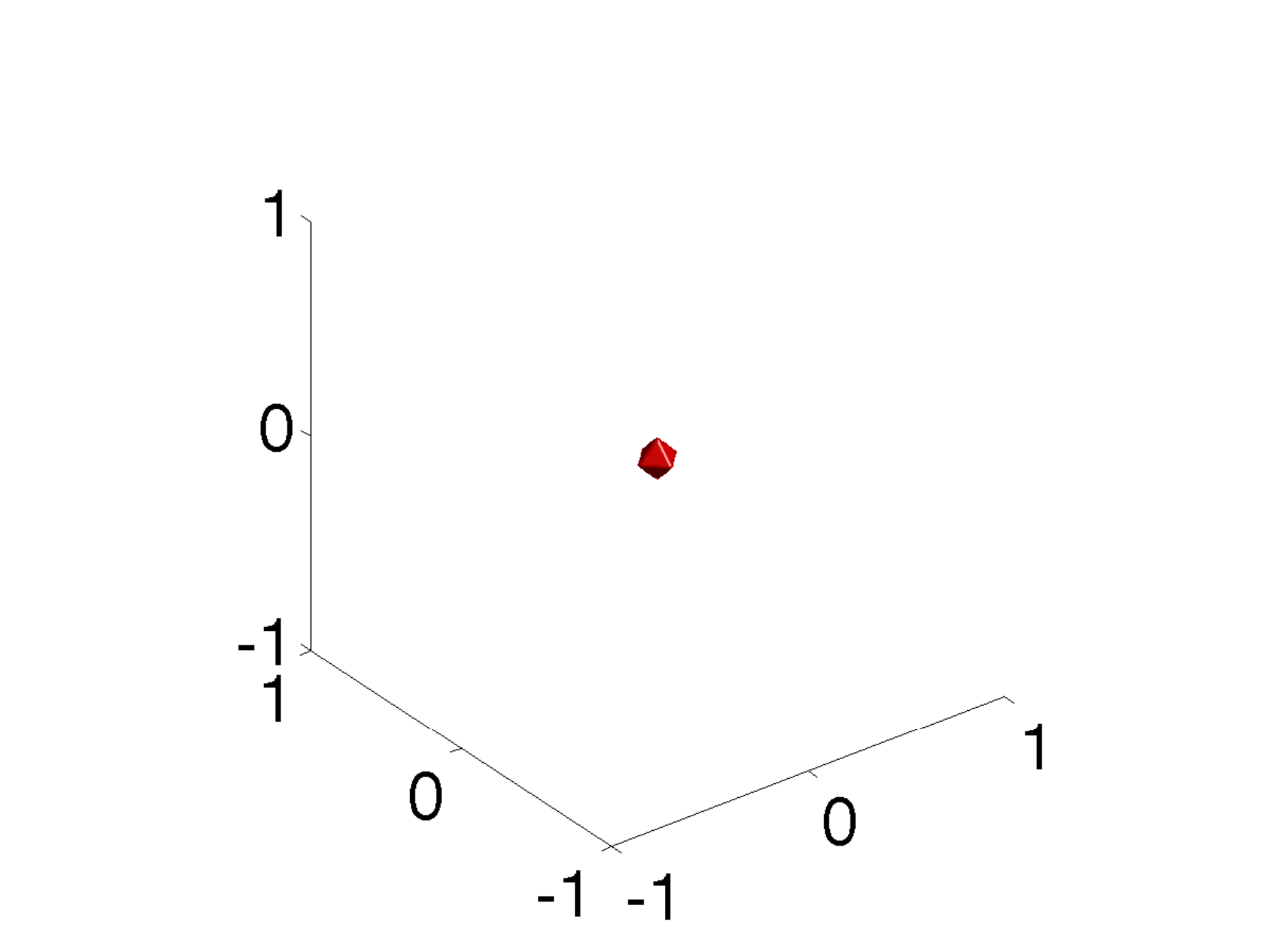}\vspace{-.2cm}\\
		\hspace{-.4cm}\includegraphics[width=0.15\textwidth, clip=true,trim=3.5cm 0cm 3.5cm 1.5cm]{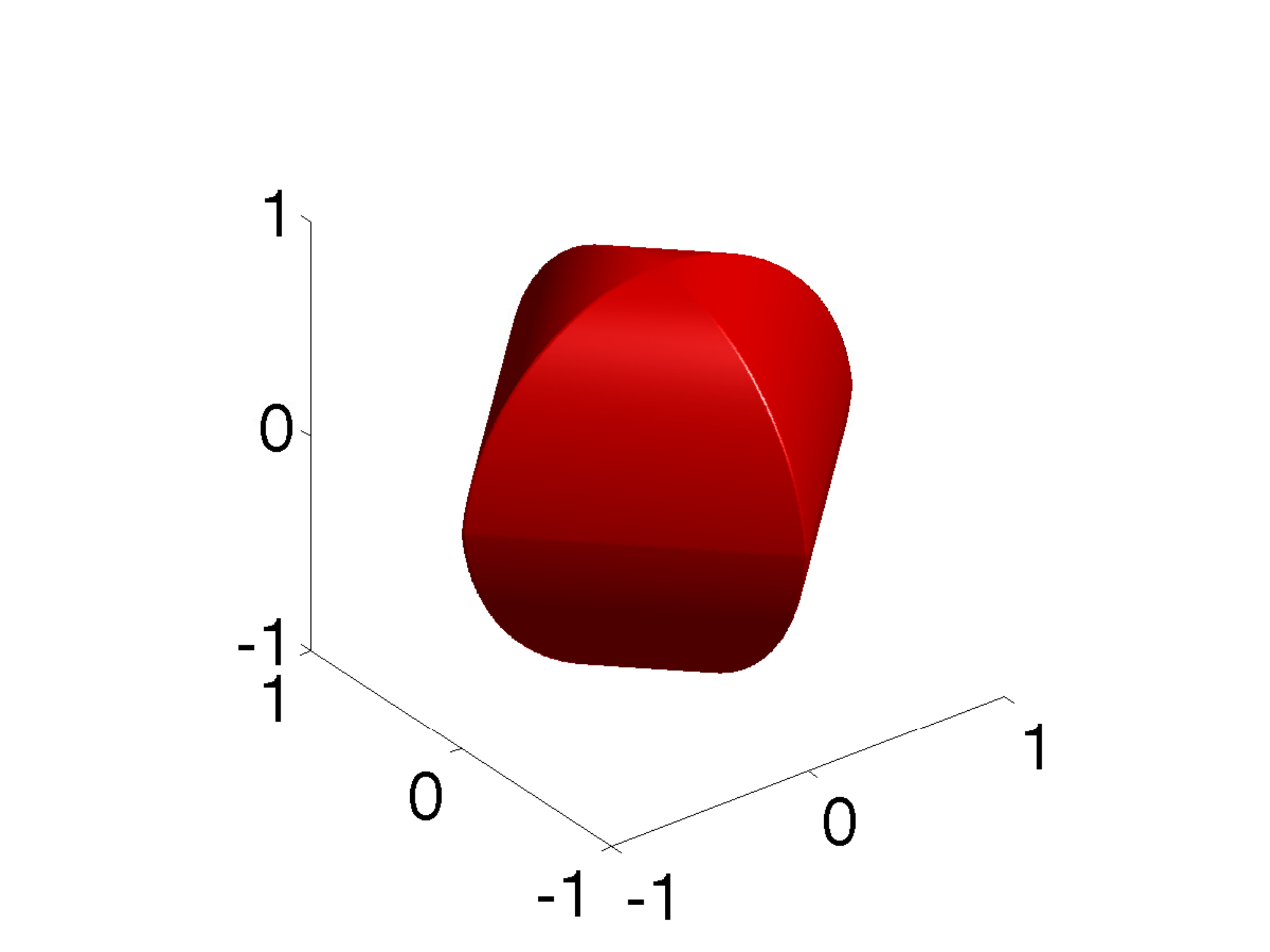} & \hspace{-.1cm}\includegraphics[width=0.15\textwidth, clip=true,trim=3.5cm 0cm 3.5cm 1.5cm]{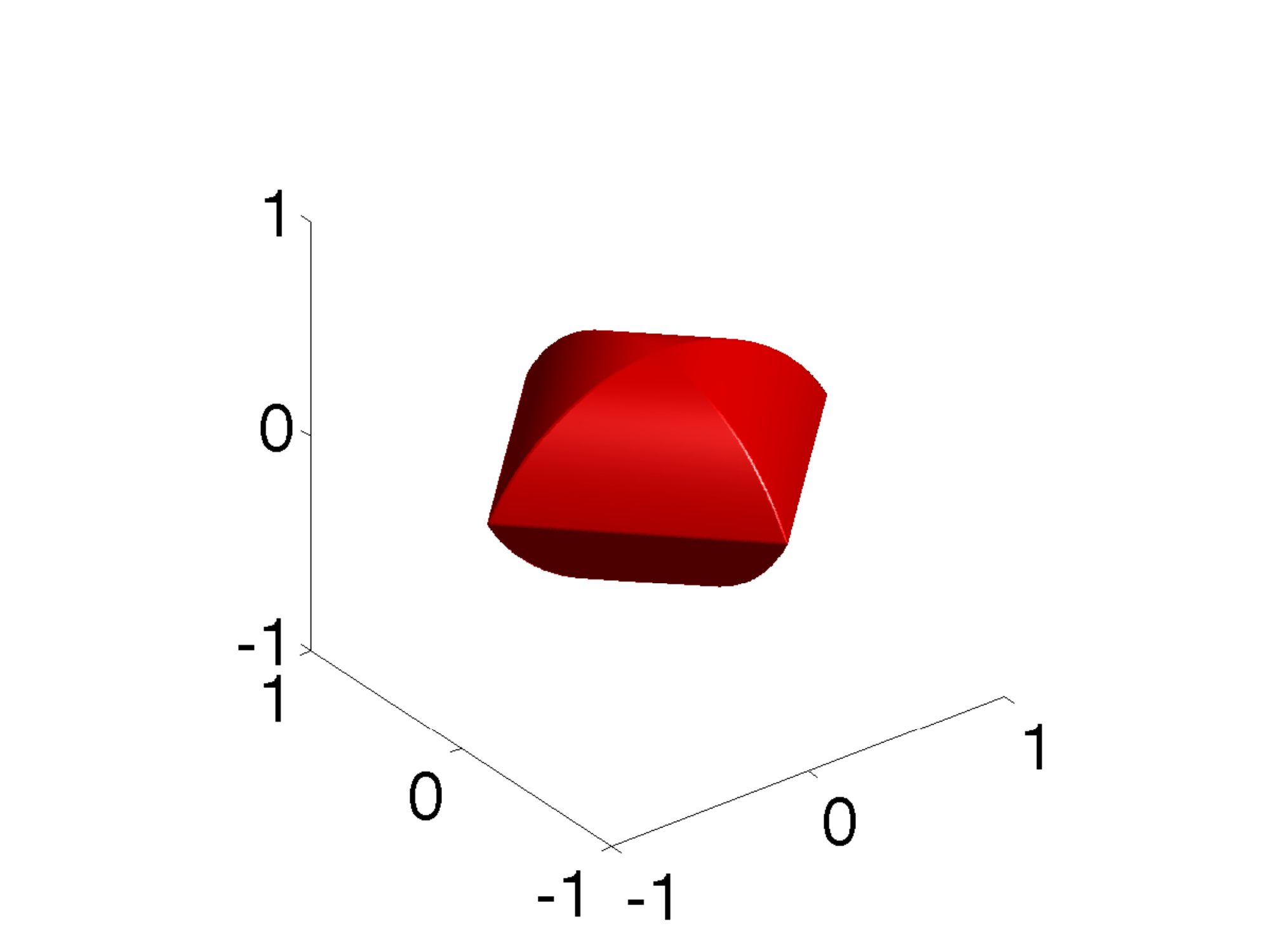} & \hspace{-.1cm}\includegraphics[width=0.15\textwidth, clip=true,trim=3.5cm 0cm 3.5cm 1.5cm]{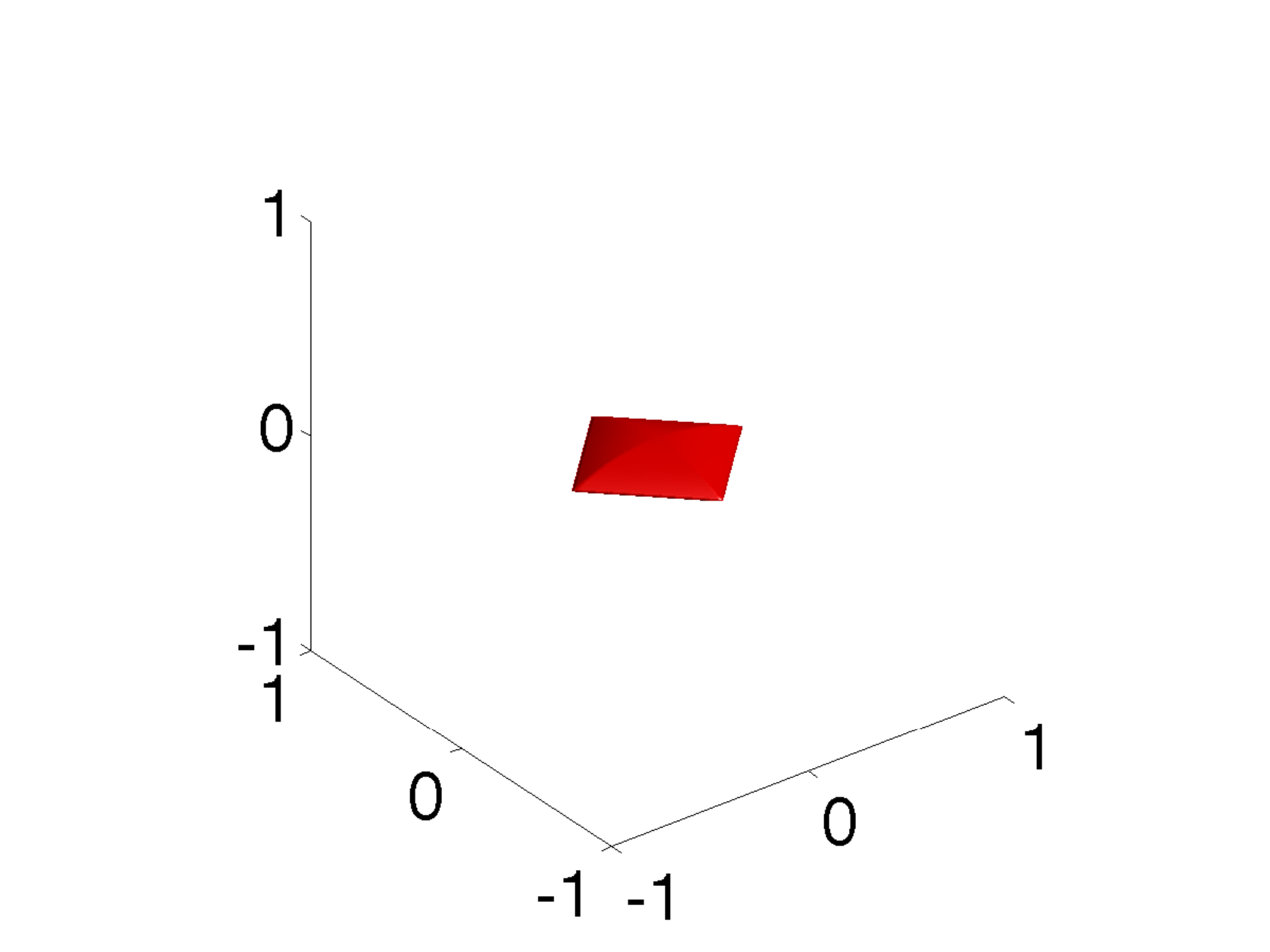}\vspace{-.2cm}\\
		\hspace{-.4cm}\includegraphics[width=0.15\textwidth, clip=true,trim=3.5cm 0cm 3.5cm 1.5cm]{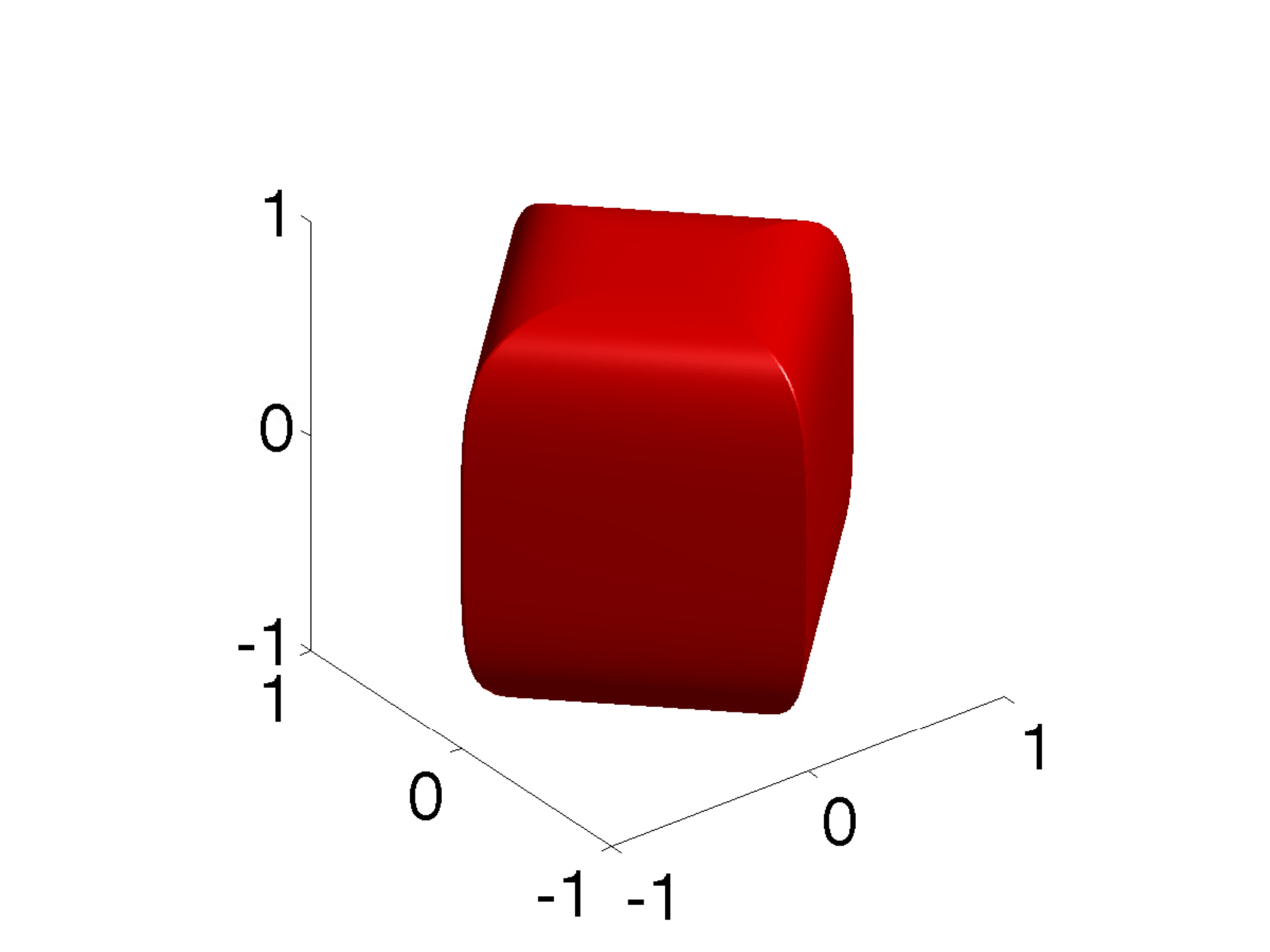} & \hspace{-.1cm}\includegraphics[width=0.15\textwidth, clip=true,trim=3.5cm 0cm 3.5cm 1.5cm]{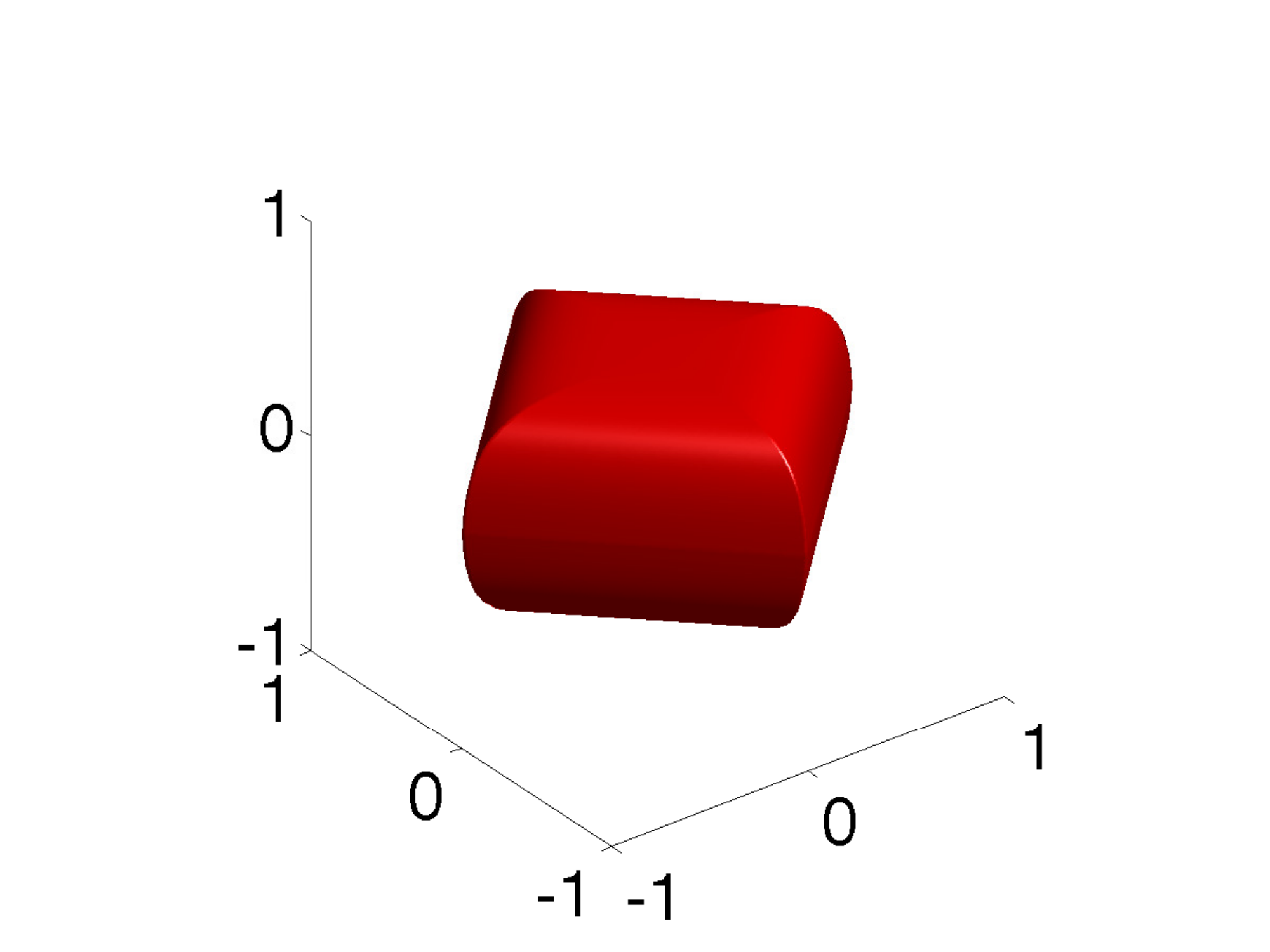} & \hspace{-.1cm}\includegraphics[width=0.15\textwidth, clip=true,trim=3.5cm 0cm 3.5cm 1.5cm]{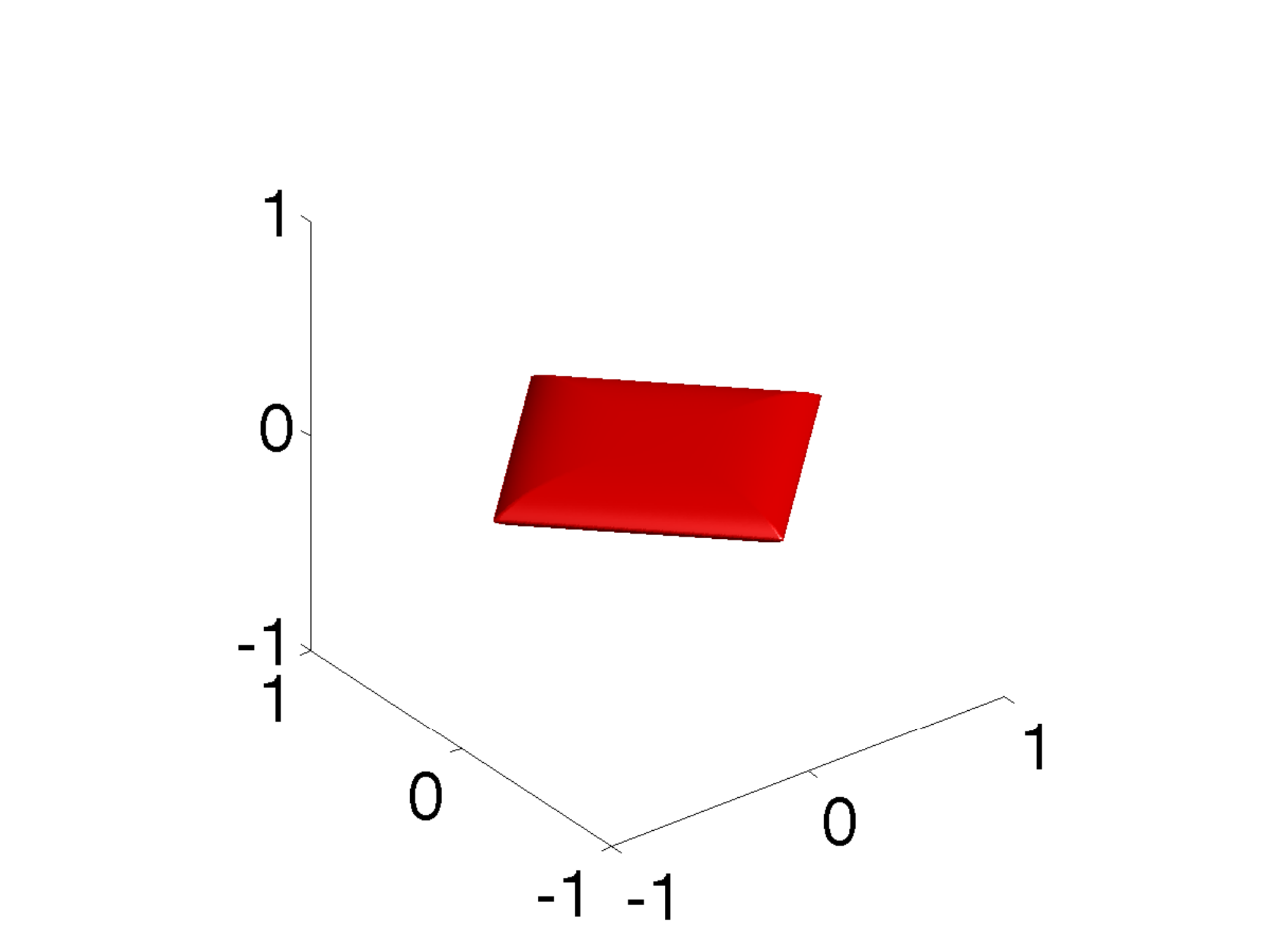}\\
		(a) $a=0.1$ &(b) $a=0.5$ & (c) $a=0.9$
	\end{tabular}
	\caption{Unit balls of the regularizer in \eqref{EqnGrpLearn1} for different values of $\alpha$. Suppose we have $\w^{(1)} = (x,y)^\top$ in group $G_1$ and $\w^{(2)} = (z,a)^\top$ in $G_2$. In order to visualize in 3-d space, we vary 3 variables $x,y$ and $z$, and fix $w^{(2)}_2$ to some constant $a$. The first row is using $\ell_1$ norm for summing the groups: $(|x|+|y|)+(|z|+a)$, the second row is using $\ell_2$ norm: $\sqrt{(|x|+|y|)^2+(|z|+a)^2}$, and the last row is using $\ell_5$ norm: $\big((|x|+|y|)^5+(|z|+a)^5\big)^{0.2}$. As $a$ increases (from the first column to the third one), $\w_1$ quickly shrinks to zero in case of $\ell_1$ summation. One the other hand, in case of $\ell_2$ summation, $x$ and $y$ in $\w^{(1)}$ are allowed to be non-zero, while $z$ shrinks to zero. This effect gets clearer as $\alpha$ increases.}
	\label{FigCompNorms}
\end{figure}

Figure \ref{FigCompNorms} visualizes the unit surfaces of different regularizers derived from \eqref{EqnGrpLearn1} (i.e. $\sum_{g \in \G}  (  \big\| \W \U_g \big\|_{1,2} )^\alpha = 1$ for different choices of $\alpha$.) for the case where we have two groups, each of which has a single task. It shows that a large norm value on one group (in this example, on $G_2$ when $a$ = 0.9) does not force the other group (i.e $G_1$) to have a small norm as $\alpha$ becomes larger. This is evidenced in the bottom two rows of the third column (compare it with how $\ell_1$ behaves in the top row). In other words, we see the benefits of using $\alpha \geq 2$ to encourage more active groups. 

While the constraint $[\U_g]_{ik} \in \{0,1\}$ in \eqref{EqnGrpLearn1} ensures hard group memberships, solving it requires integer programming which is intractable in general. Therefore, we relax the constraint on $\U$ to $\quad 0 \leq [\U_g]_{ik} \leq 1$.
However, this relaxation along with the $\ell_\alpha$ norm over groups prevents both $\|\W\U_g\|_{1,2}$ and also individual $[\U_g]_{ik}$ from being zero. For example, suppose that we have two tasks (in $\reals^{2}$) in a single group, and $\alpha =2$. Then, the regularizer for any $g$ can be written as $\Big( \sqrt{ (w^{(1)}_1)^2 \ u_{g,1}^2 + (w^{(2)}_1)^2 \ u_{g,2}^2} + \sqrt{ (w^{(1)}_2)^2\ u_{g,1}^2 + (w^{(2)}_2)^2\ u_{g,2}^2 }\Big)^2$. To simply the situation, assume further that all entries of $\W$ are uniformly a constant $w$. Then, this regularizer for a single $g$ would be simply reduced to $4 w^2 (u_{g,1}^2 + u_{g,2}^2)$, and therefore the regularizer over all groups would be $4 w^2 ( \sum_{t=1}^m \sum_{g} u_{g,t}^2 )$. Now it is clearly seen that the regularizer has an effect of grouping over the group membership vector $(u_{g_1,t}, u_{g_2,t},\hdots, u_{g_{N},t})$ and encouraging the set of membership parameters for each task to be uniform.
 
 To alleviate this challenge, we re-parameterize $u_{g,t}$ with a new membership parameter $u'_{g,t} := \sqrt{u_{g,t}}$. The constraint does not change with this re-parameterization: $0 \leq u'_{g,t} \leq 1$. Then, in the previous example, the regularization over all groups would be (with some constant factor) the sum of $\ell_1$ norms, $\| (u_{g_1,t}, u_{g_2,t},\hdots, u_{g_{N},t}) \|_1$ over all tasks, which forces them to be sparse. Note that even with this change, the activations of groups are not sparse since the sum over groups is still done by the $\ell_2$ norm. 

\begin{figure}[t]
	\centering
	\begin{tabular}{cccc}
	\subfigure[$\W$ for 30 tasks]{\includegraphics[width=0.25\textwidth]{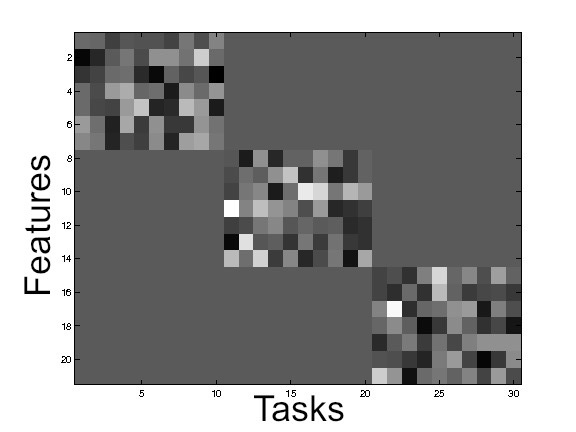}} 
	& 
		\subfigure[Learned $\U$ from \eqref{EqnBaseline}]{\includegraphics[width=0.23\textwidth]{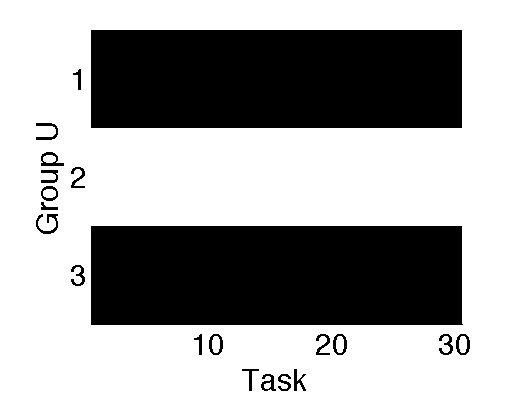}}
		&
   \subfigure[Learned $\U$ from \eqref{EqnBaseline} after relaxing the integer constraints on $\U$]{\includegraphics[width=0.25\textwidth]{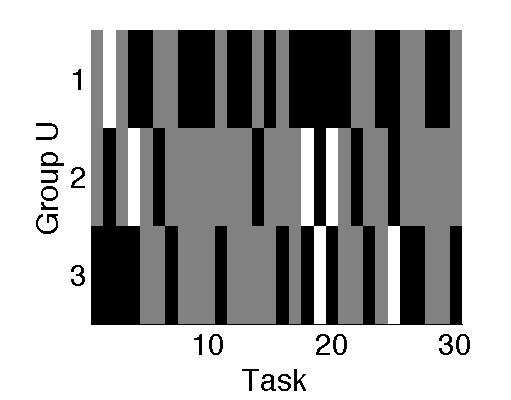}}
    & 
    \subfigure[Learned $\U$ from \eqref{EqnGrpLearn3}]{\includegraphics[width=0.25\textwidth]{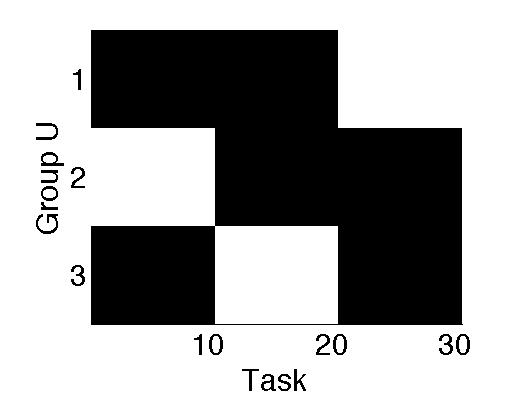}} 
	\end{tabular}
	\caption{Comparisons on different regularizers. (a) $\W$ for 30 tasks with 21 features, which is assumed to be \emph{known and fixed}. Three groups are clearly separated: ($T_1$-$T_{10}$), ($T_{11}$-$T_{20}$), ($T_{21}$-$T_{30}$) whose nonzero elements are block diagonally located (black is negative, white is positive). (b) Learned $\U$ from \eqref{EqnBaseline} (with a relaxing of discrete constraints, and the square root reparameterization). All tasks are assigned to a single group. (c) Estimation on $\U$ using \eqref{EqnBaseline} after relaxing the integer constraints on $U$. All there groups are active (have tasks in them), but most of $u_{g,t}$ are not binary. (d) Estimation on $\U$ using \eqref{EqnGrpLearn3}. All group memberships are recovered correctly. Note that the order of groups does not matter. For (b)-(d), white is 1 and black is 0.}
	\label{FigToyComp}
\end{figure}

\begin{figure}
	\centering
		\begin{tabular}{cc}
		\hspace{-.45cm}\subfigure[$Q$]{\includegraphics[width=0.25\textwidth]{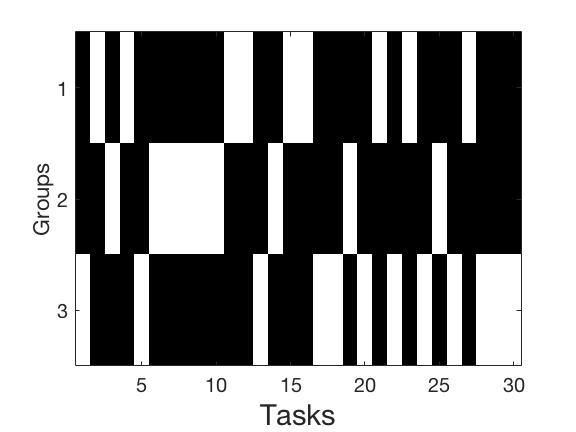}} & 		
		\hspace{-.4cm}\subfigure[$\W$]{\includegraphics[width=0.25\textwidth]{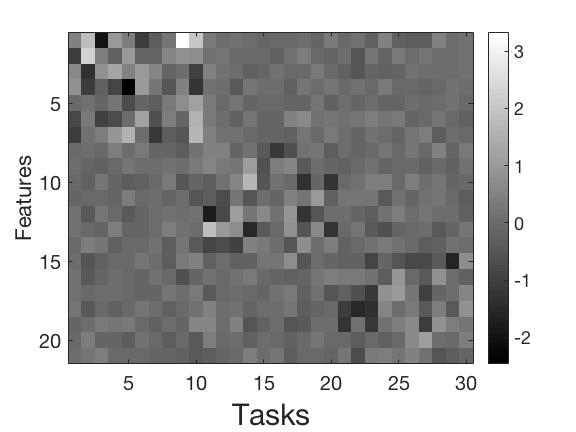}} \\
		Task-grouping & Weights matrix
	\end{tabular}
	\caption{Comparison of the regularizer from \citet{kang2011} when learned on our synthetic dataset (set-1). Fig \ref{FigToyComp} (a) shows the $\W^*$. (a) the learned $Q$, analogous to $\U$ in our notation (white is 1 and black is 0) and (b) $\widehat{W}$}
	\label{FigToyKang}
\end{figure}

Toward this, we finally introduce the following problem to jointly estimate $\U$ and $\W$ (specifically with focus on the case when $\alpha$ is set to $2$):
\begin{align}\label{EqnGrpLearn3}
	\minimize_{\W,\U} & \, \sum_{t=1}^{m}  \L(\w^{(t)}; \mathcal{D}_t)   + \sum_{g \in \G} \lambda_g \Big( \big\| \W \sqrt{\U_g} \big\|_{1,2} \Big)^2  \nonumber\\
	\st & \, \sum_{g \in \G} \U_g = \I^{m\times m} \ , \quad 0 \leq [\U_g]_{ik} \leq 1\, .
\end{align}
where $\sqrt{\bm M}$ for a matrix $\bm M$ is obtained from element-wise square root operations of $\bm M$.   
Note that \eqref{EqnGrpLearn1} and \eqref{EqnGrpLearn3} are not equivalent, but the minimizer $\widehat{\U}$ given any fixed $\W$ usually is actually binary or has the same objective with some other binary $\widehat{\U}'$ (see Theorem 1 of \citet{kang2011} for details). As a toy example, we show in Figure \ref{FigToyComp}, the estimated $\U$ (for a known $\W$) via different problems presented so far (\eqref{EqnBaseline}, \eqref{EqnGrpLearn3}).

\paragraph{Fusing the Group Assignments}
\label{sec:fusion}
The approach derived so far works well when the number of groups $N << m$, but can create many singleton groups when $N$ is very large. We add a final modification to our objective to encourage tasks to have similar group membership wherever warranted. This makes the method more robust to the mis-specification of the number of groups, `$N$' as it prevents the grouping from becoming too fragmented when $N >> N^*$. 
For each task $t=1,\hdots,m$, we define $N \times N$ matrix $\V_t := \diag(u_{1,t},\hdots,u_{N,t}).$ Note that the $\V_t$ are entirely determined by the $\U_g$ matrices, so no actual additional variables are introduced.
Equipped with this additional notation, we obtain the following objective
where $\|\cdot\|_\F$ denotes the Frobenius norm (the element-wise $\ell_2$ norm), and $\mu$ is the additional regularization parameter that controls the number of active groups. 
\begin{align}\label{EqnGrpLearnFus}
	\minimize_{\W,\U} & \, \sum_{t=1}^{m}  \L(\w^{(t)}; \mathcal{D}_t)   + \sum_{g \in \G} \lambda_g \Big( \big\| \W \sqrt{\U_g} \big\|_{1,2} \Big)^2  \nonumber\\
	& +  \mu \sum_{t<t'} \big\| \V_t - \V_{t'} \big\|_{\F}^2 \nonumber\\
	\st & \, \sum_{g \in \G} \U_g = \I^{m\times m} \ , \quad [\U_g]_{ik} \in [0,1] \, 
\end{align}

%
%

\subsection{Theoretical comparison of approaches:}
\label{sec:complexity}
It is natural to ask whether enforcing the shared sparsity structure,
when groups are unknown leads to any efficiency in the number of
samples required for learning. In this section, we will use intuitions
from the high-dimensional statistics literature in order to compare
the sample requirements of different alternatives such as independent
lasso or the approach of ~\citet{kang2011}. Since the formal analysis of
each method requires making different assumptions on the data $X$ and
the noise, we will instead stay intentionally informal in this
section, and contrast the number of samples each approach would
require, assuming that the desired structural conditions on the $x$'s
are met. We evaluate all the methods under an idealized setting where
the structural assumptions on the parameter matrix $W$ motivating our
objective~\eqref{EqnBaseline} hold exactly. That is, the parameters
form $N$ groups, with the weights in each group taking non-zero values
only on a common subset of features of size at most $s$. We begin with
our approach.

\paragraph{Complexity of Sparsity Grouped MTL} 
Let us consider the simplest inefficient version of our method, a
generalization of subset selection for Lasso which searches over all
feature subsets of size $s$. It picks one subset $S_g$ for each group
$g$ and then estimates the weights on $S_g$ independently for each
task in group $g$. By a simple union bound argument, we expect this
method to find the right support sets, as well as good parameter
values in $O(N s\log d + ms)$ samples. This is the complexity of
selecting the right subset out of $d \choose s$ possibilities for
each group, followed by the estimation of $s$ weights for each
task. We note that there is no direct interaction between $m$ and $d$
in this bound.
\paragraph{Complexity of independent lasso per task} An alternative
approach is to estimate an $s$-sparse parameter vector for each task
independently. Using standard bounds for $\ell_1$ regularization (or
subset selection), this requires $O(s\log d)$ samples per task,
meaning $O(ms\log d)$ samples overall. We note the multiplicative
interaction between $m$ and $\log d$ here.
\paragraph{Complexity of learning all tasks jointly} A different
extreme would be to put all the tasks in one group, and enforce shared
sparsity structure across them using $\|\cdot\|_{1,2}$ regularization
on the entire weight matrix. The complexity of this approach depends
on the sparsity of the union of all tasks which is $Ns$, much larger
than the sparsity of individual groups. Since each task requires to
estimate its own parameters on this shared sparse basis, we end up
requiring $O(msN\log d)$ samples, with a large penalty for ignoring the
group structure entirely. 
\paragraph{Complexity of \citet{kang2011}} As yet another baseline, we
observe that an $s$-sparse weight matrix is also naturally low-rank
with rank at most $s$. Consequently, the weight matrix for each group
has rank at most $s$, plausibly making this setting a good fit for the
approach of \citet{kang2011}. However, appealing to the standard results
for low-rank matrix estimation (see e.g. \citet{negahban2011}), learning a
$d\times n_g$ weight matrix of rank at most $s$ requires $O(s(n_g +
d))$ samples, where $n_g$ is the number of tasks in the group
$g$. Adding up across tasks, we find that this approach requires a
total of $O(s(m + md))$, considerably higher than all other baselines
even if the groups are already provided. It is easy to see why this is
unavoidable too. Given a group, one requires $O(ms)$ samples to
estimate the entries of the $s$ linearly independent rows. A method
utilizing sparsity information knows that the rest of the columns are
filled with zeros, but one that only knows that the matrix is low-rank
assumes that the remaining $(d-s)$ rows all lie in the linear span of
these $s$ rows, and the coefficients of that linear combination need
to be estimated giving rise to the additional sample complexity. In a
nutshell, this conveys that estimating a sparse matrix using low-rank
regularizers is sample inefficient, an observation hardly surprising
from the available results in high-dimensional statistics but
important in comparison with the baseline of~\citet{kang2011}. 

For ease of reference, we collect all these results in
Table~\ref{table:complexities} below.
\begin{table}[th]
  \begin{tabular}{|c|c|c|c|c|}
    \hline
    & \textbf{SG-MTL} & Lasso & Single group & \citet{kang2011}\\\hline 
    Samples & $O(N s\log d + ms)$ & $O(ms\log d)$ &
    $O(msN\log d)$ & $O(s(m + md))$ \\\hline
  \end{tabular}
  \caption{Sample complexity estimates of recovering group memberships
  and weights using different approaches}
  \label{table:complexities}
\end{table}

\vspace{-1cm}
\section{Optimization}
\begin{algorithm}[t]
	\caption{\AlgoSize SG-MTL (eqn \eqref{EqnGrpLearn3})}
	\label{AlgAlt}
	\begin{algorithmic}
		\AlgoSize
		\STATE {\bfseries Input:} $\{\mathcal{D}_t\}_{t=1}^m$
		\STATE Initialize $\W$ using single task learning
		\STATE Initialize $\U$ to random matrix
		\REPEAT
			\STATE Update $\U$ by solving a projected gradient descent
			\FOR {all tasks $t = 1,2,\hdots, m$}
				\STATE Update $\w^{(t)}$ using a coordinate descent for all features $j=1,2,\hdots,d$
			\ENDFOR
		\UNTIL{stopping criterion is satisfied}
	\end{algorithmic}
\end{algorithm}

We solve \eqref{EqnGrpLearn3} by alternating minimization: repeatedly solve one variable fixing the other until convergence (Algorithm \ref{AlgAlt}) We discuss details below.\\
\noindent{\textbf{Solving \eqref{EqnGrpLearn3} w.r.t  $\U$}}: This step is challenging since we lose convexity due the reparameterization with a square root. The solver might stop with a premature $\U$ stuck in a local optimum. However, in practice, we can utilize the random search technique to get the minimum value over multiple re-trials. Our experimental results reveal that the following projected gradient descent method performs well.  
 
Given a fixed $\W$, solving for $\U$ only involves the regularization term i.e $R(\U)= \sum_{g \in \G} \lambda_g \Big( \sum_{j=1}^d   \big \| \W_j \sqrt{\U_g} \big \|_2  \Big)^2$ which is differentiable w.r.t $U$. The derivative is shown in the appendix along with the extension for the fusion penalty from \eqref{EqnGrpLearnFus}.
Finally after the gradient descent step, we project $(u_{g_1,t}, u_{g_2,t}, \hdots, u_{g_{N},t})$ onto the simplex (independently repeat the projection for each task) to satisfy the constraints on it. Note that projecting a vector onto a simplex can be done in $O(m \log m)$ \citep{Chen2011}.

\paragraph{Solving \eqref{EqnGrpLearn3} w.r.t  $\W$}
This step is more amenable in the sense that \eqref{EqnGrpLearn3} is convex in $\W$ given $\U$. However, it is not trivial to efficiently handle the complicated regularization terms. Contrast to $\U$ which is bounded by $[0,1]$, $\W$ is usually unbounded which is problematic since the regularizer is (not always, but under some complicated conditions discovered below) non-differentiable at $0$.  

While it is challenging to directly solve with respect to the entire $\W$,  we found out that the coordinate descent (in terms of each element in $\W$) has a particularly simple structured regularization.

Consider any $w_j^{(t)}$ fixing all others in $\W$ and $\U$; the regularizer $R(\U)$ from \eqref{EqnGrpLearn3} can be written as
\begin{align}\label{EqnCoordW}
	& \sum_{g \in \G} \lambda_g \Bigg\{ u_{g,t} (w_j^{(t)})^2 +  \Bigg.  2 \bigg(\sum_{j' \neq j} \sqrt{\sum_{t'=1}^m  u_{g,t'} (w_{j'}^{(t')})^2 } \bigg) \sqrt{\sum_{t'=1}^m  u_{g,t'} (w_j^{(t')})^2 } \Bigg. \Bigg\} + C(j,t)
\end{align}
where $w_j^{(t)}$ is the only variable in the optimization problem, and $C(j,t)$ is the sum of other terms in \eqref{EqnGrpLearn3} that are constants with respect to $w_j^{(t)}$.

For notational simplicity, we define $\kappa_{g,t} :=  \sum_{t' \neq t}  u_{g,t'} (w_j^{(t')})^2$ that is considered as a constant in \eqref{EqnCoordW} given $\U$ and $\W\setminus \{w_j^{(t)}\}$. Given $\kappa_{g,t}$ for all $g \in \G$, we also define $\G^0$ as the set of groups such that $\kappa_{g,t} = 0$ and $\G^+$ for groups s.t. $\kappa_{g,t} >0$. Armed with this notation and with the fact that $\sqrt{x^2} = |x|$, we are able to rewrite \eqref{EqnCoordW} as
\begin{align}\label{EqnCoordW2}
	& \sum_{g \in \G} \lambda_g  u_{g,t} (w_j^{(t)})^2 + 2\sum_{g \in \G^+} \lambda_g \bigg(\sum_{j' \neq j} \sqrt{\sum_{t'=1}^m  u_{g,t'} (w_{j'}^{(t')})^2 } \bigg) \sqrt{\sum_{t'=1}^m  u_{g,t'} (w_j^{(t')})^2 } \\
	& + 2\sum_{g \in \G^0} \lambda_g \bigg(\sum_{j' \neq j} \sqrt{\sum_{t'=1}^m  u_{g,t'} (w_{j'}^{(t')})^2 } \bigg) \sqrt{u_{g,t}} \big| w_j^{(t)} \big| \nonumber
\end{align}
where we suppress the constant term $C(j,t)$. Since $\sqrt{x^2 + a}$ is differentiable in $x$ for any constant $a > 0$, the first two terms in \eqref{EqnCoordW2} are differentiable with respect to $w_j^{(t)}$, and the only non-differentiable term involves the absolute value of the variable, $| w_j^{(t)}|$. As a result, \eqref{EqnCoordW2} can be efficiently solved by proximal gradient descent followed by an element-wise soft thresholding. Please see appendix for the gradient computation of $\L$ and soft-thresholding details.

\section{Experiments}

We conduct experiments on two synthetic and two real datasets and compare with the following approaches. \\
1) Single task learning (STL): Independent models for each task using elastic-net regression/classification. \\
2) AllTasks: We combine data from all tasks into a single task and learn an elastic-net model on it\\
3) Clus-MTL: We first learn STL for each task, and then cluster the task parameters using k-means clustering. For each task cluster, we then train a multitask lasso model. \\
4) GO-MTL: group-overlap MTL \citep{kumar2012} \\
5) \citet{kang2011}: nuclear norm based task grouping \\
6) SG-MTL : our approach from equation \ref{EqnGrpLearn3} \\
7) Fusion SG-MTL: our model with a fusion penalty (see Section \ref{sec:fusion})

\begin{table}[h]
\caption{Synthetic datasets: (Upper table) Average MSE from 5 fold CV. (Lower table) Varying group sizes and the corresponding average MSE. For each method, lowest MSE is highlighted.}
\label{synth_results}
\setlength{\tabcolsep}{3pt}
{\small
\centering
\begin{tabular}{c|rrrrr} 
\toprule
Dataset & STL & ClusMTL & Kang & SG-MTL & Fusion SG-MTL \\ \midrule
set-1 (3 groups) & 1.067 & 1.221 & 1.177 & 0.682 & 0.614 \\
set-2 (5 groups) & 1.004 & 1.825 & 0.729 & 0.136 & 0.130 \\
\bottomrule
\multicolumn{6}{c}{} \\
\multicolumn{6}{c}{\textbf{Synthetic data-2 with 30\% feature overlap across groups}} \\ \toprule
 & \multicolumn{5}{c}{\textit{Number of groups}, $N$} \\ 
\textit{Method} & 2 & 4 & $N^*$=5 & 6 & 10 \\ \midrule
ClusMTL & 1.900 & 1.857 & 1.825 & 1.819 & \textbf{1.576} \\
\citet{kang2011} & \textbf{0.156} & 0.634 & 0.729 & 0.958 & 1.289 \\
SG-MTL & 0.145 & \textbf{0.135} & 0.136 & 0.137 & 0.137 \\
Fusion SG-MTL & 0.142 & 0.139 & \textbf{0.130} & 0.137 & 0.137 \\
\bottomrule
\end{tabular}
}
\end{table}

\subsection{Results on synthetic data}
\label{sec:synthdata}
\begin{figure}
\centering
\includegraphics[scale=0.06]{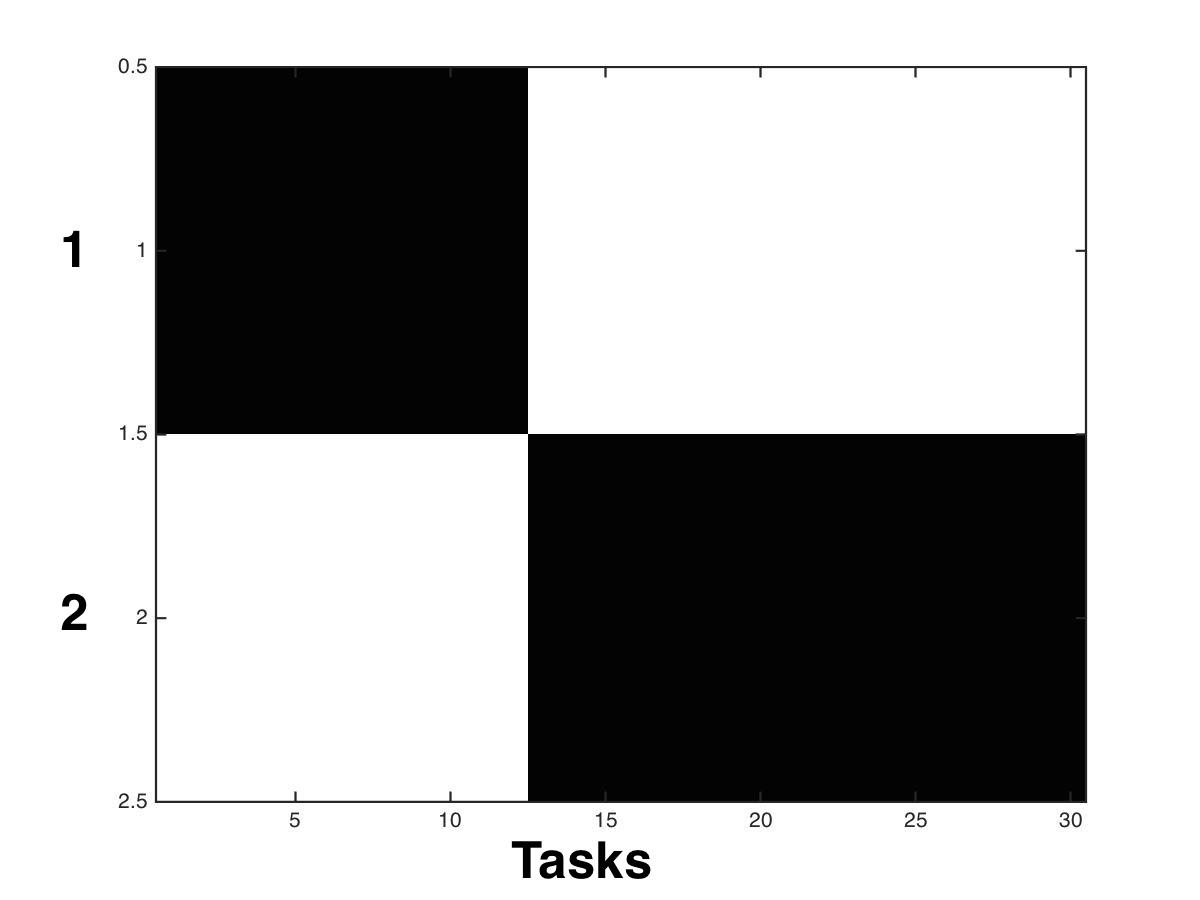}
\includegraphics[scale=0.06]{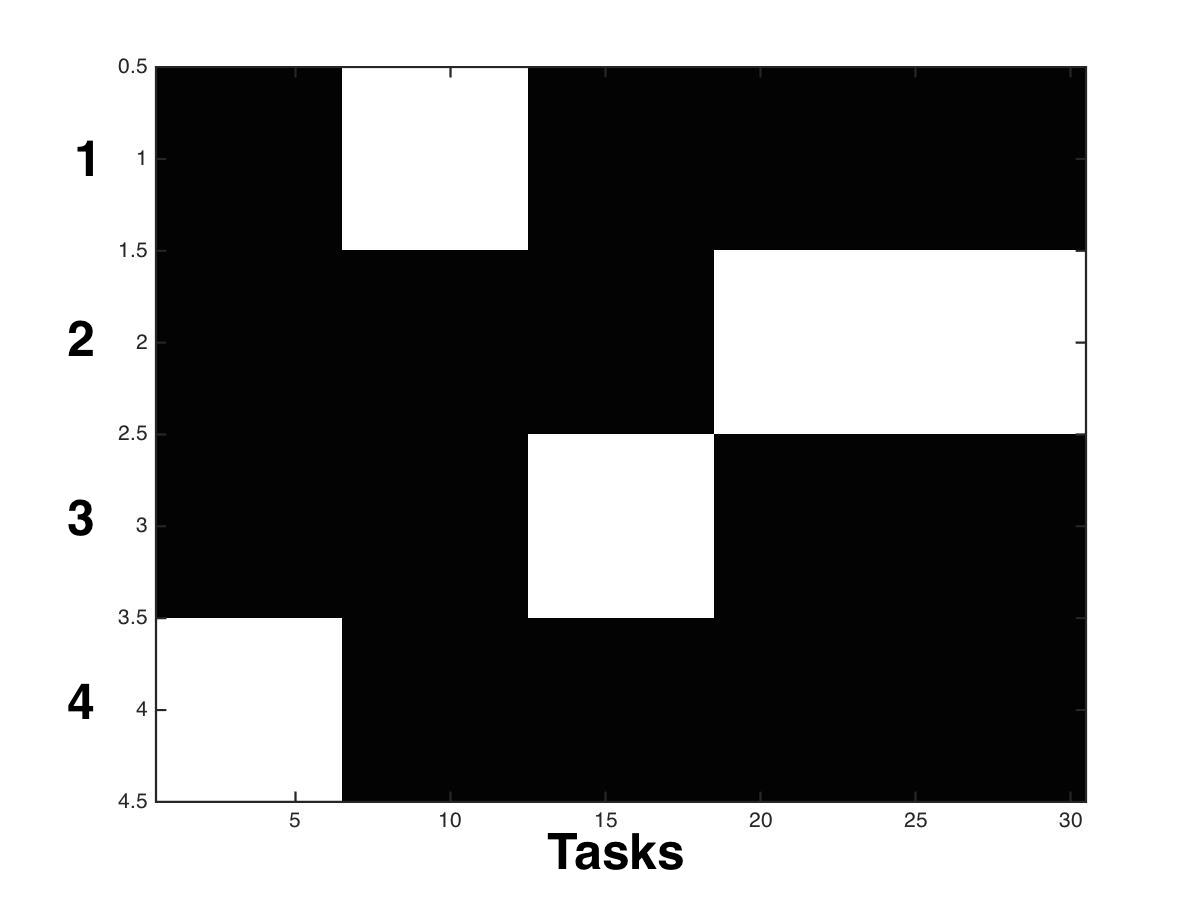}
\includegraphics[scale=0.06]{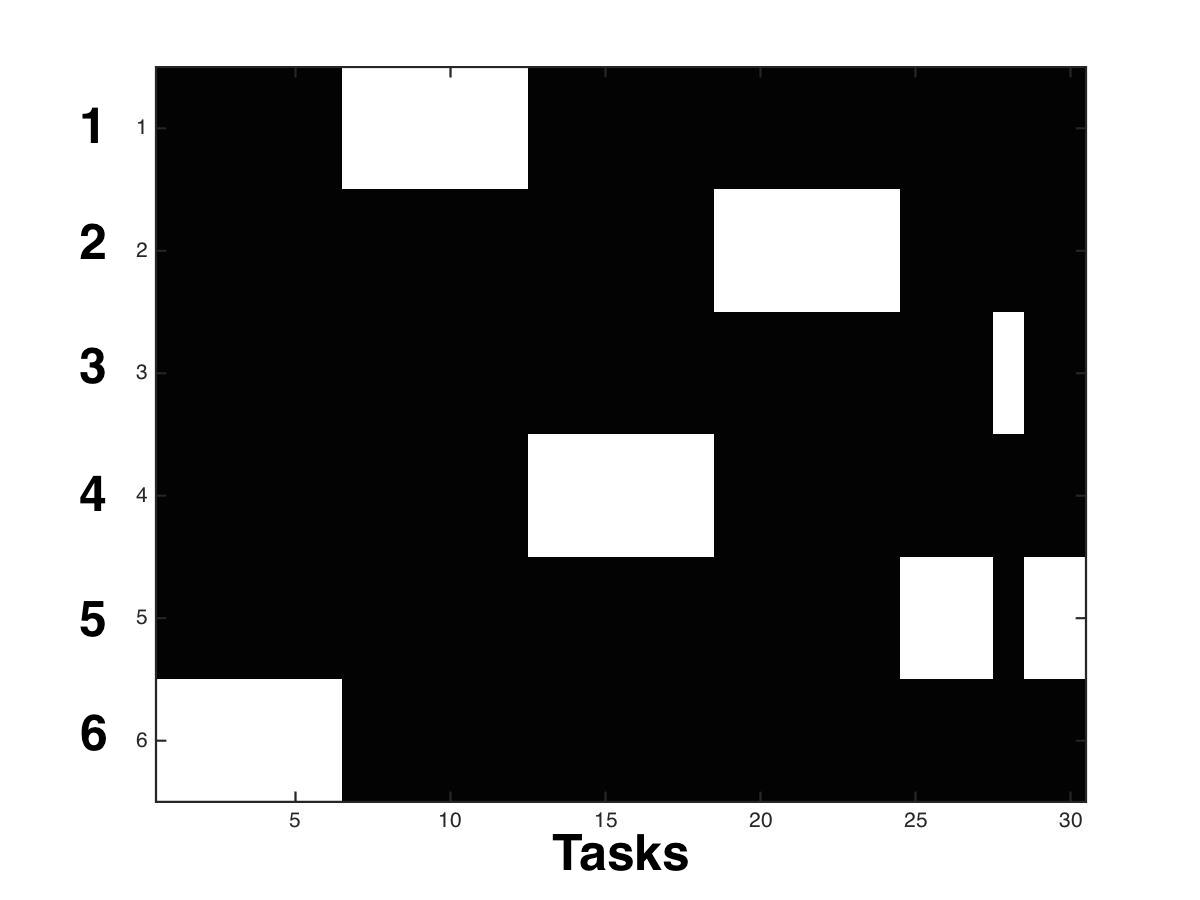}
\caption{Groupings found for the case where $N^*$=5. We show results of a typical run of our method with $N$=2,4,6. On the x-axis are the 30 tasks and on the y-axis are the group ids.}
\label{fig:groups}
\end{figure}

The first setting is similar to the synthetic data settings used in \citet{kang2011} except for how $W$ is generated (see Fig \ref{FigToyComp}(a) for our parameter matrix and compare it with Sec 4.1 of \citet{kang2011}). We have 30 tasks forming 3 groups with 21 features and 15 examples per task. Each group in $\W$ is generated by first fixing the zero components and then setting the non-zero parts to a random vector $w$ with unit variance. $Y_t$ for task $t$ is $X_t W_t + \epsilon$. For the second dataset, we generate parameters in a similar manner as above, but with 30 tasks forming 5 groups, 100 examples per task, 150 features and a 30\% overlap in the features across groups. In table \ref{synth_results}, we show 5-fold CV results and in Fig \ref{fig:groups} we show the groups ($\U$) found by our method. 

\noindent\textbf{How many groups?}
Table \ref{synth_results} (Lower) shows the effect of increasing group size on three methods (smallest MSE is highlighted). For our methods, we observe a dip in MSE when $N$ is close to $N^*$. In particular, our method with the fusion penalty gets the lowest MSE at $N^*=5$. Interestingly, \citet{kang2011} seems to prefer the smallest number of clusters, possibly due to the low-rank structural assumption of their approach, and hence cannot be used to learn the number of clusters in a sparsity based setting.

\begin{table}[h]
\begin{minipage}{0.51\textwidth}
\centering
\setlength{\tabcolsep}{4pt}
{\small
\begin{tabular}{c|rrrr} 
\toprule
& $\mu_{MSE}$ & $\sigma_{mse}$ & $\mu_{R^2}$ & $\sigma_{R2}$ \\ \midrule
STL & 0.811 &  0.02 & 0.223 & 0.01 \\
AllTasks  & 0.908 & 0.01 & 0.092 & 0.01 \\
ClusMTL & 0.823 & 0.02 & 0.215 & 0.02 \\
GOMTL & 0.794 & 0.01 & 0.218 & 0.01 \\
Kang & 1.011 & 0.03 & 0.051 & 0.03 \\
Fusion SG-MTL & \textbf{0.752} & 0.01 & \textbf{0.265} & 0.01 \\
\bottomrule
\end{tabular}
\caption{QSAR prediction: average MSE and $R^2$ over 10 train:test splits with 100 examples per task in the training split (i.e $n$=100). The standard deviation of MSE is also shown. For all group learning methods, we use $N$=5.}
\label{qsar_results}
}
\end{minipage} \hfill
\begin{minipage}{0.46\textwidth}
\centering
\includegraphics[scale=0.3]{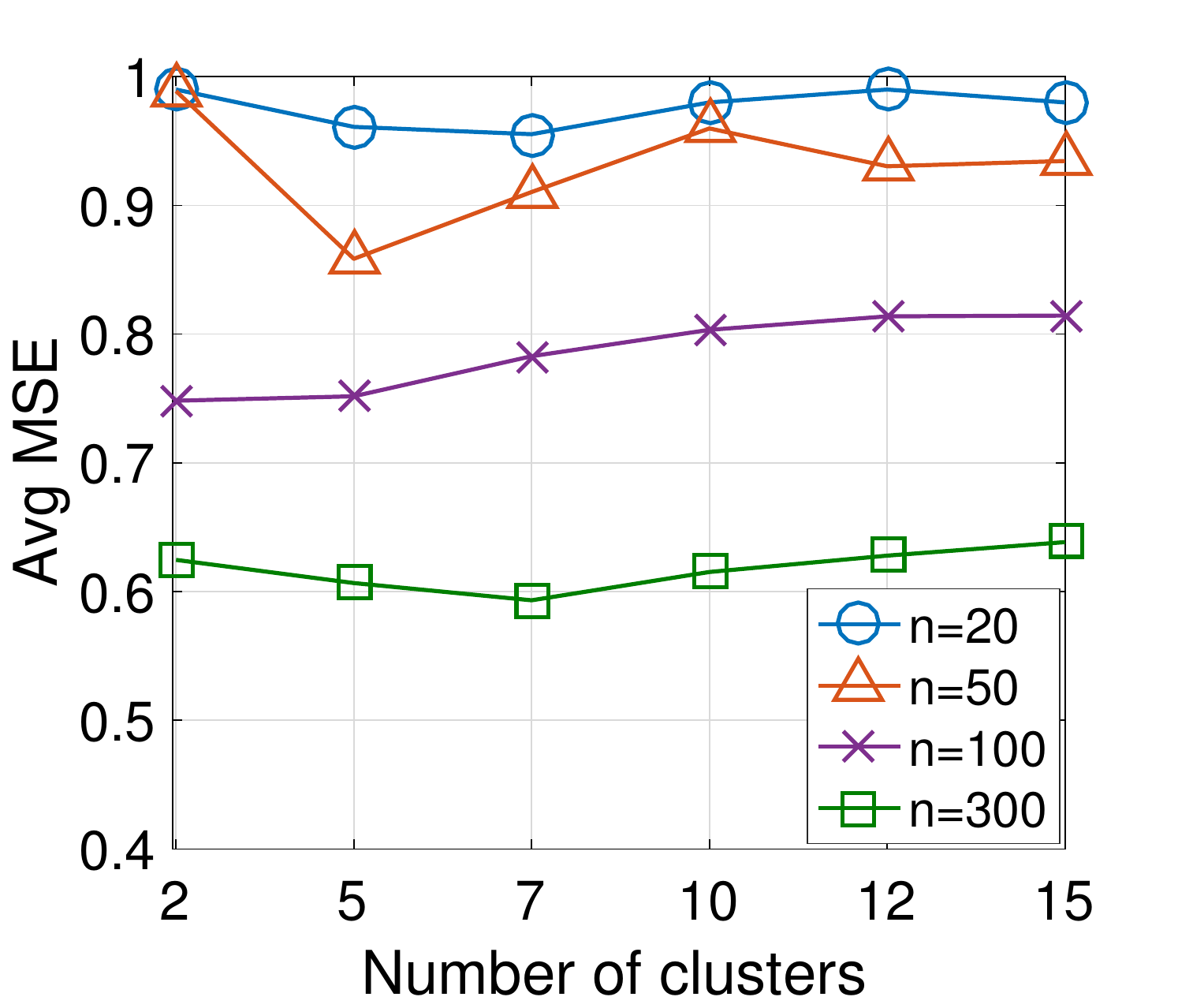}
\caption{Avg. MSE of our method (Fusion SG-MTL) as a function of the number of clusters and training data size. The best average MSE is observed with 7 clusters and $n$=300 training examples (green curve with squares). The corresponding $R^2$ for this setting is 0.401.}
\label{fig:learning_curve}
\end{minipage}
\end{table}

\subsection{Quantitative Structure Activity Relationships (QSAR) Prediction: Merck dataset}
Given features generated from the chemical structures of candidate drugs, the goal is to predict their molecular activity (a real number) with the target. This dataset from Kaggle consists of 15 molecular activity data sets, each corresponding to a different target, giving us 15 tasks. There are between 1500 to 40000 examples and $\approx$ 5000 features per task, out of which 3000 features are common to all tasks. We create 10 train:test splits with 100 examples in training set (to represent a setting where $n \ll d$) and remaining in the test set. We report $R^2$ and MSE aggregated over these experiments in Table \ref{qsar_results}, with the number of task clusters $N$ set to 5 (for the baseline methods we tried $N$=2,5,7). We found that Clus-MTL tends to put all tasks in the same cluster for any value of $N$. Our method has the lowest average MSE.

In Figure \ref{fig:learning_curve}, for our method we show how MSE changes with the number of groups $N$ (along x-axis) and over different sizes of the training/test split. The dip in MSE for $n=50$ training examples (purple curve marked `x') around $N=5$ suggests there are 5 groups. The learned groups are shown in the appendix in Fig \ref{tasks_drugs} followed by a discussion of the groupings.


\begin{table}[t]
\begin{minipage}{0.35\textwidth}
\centering
\setlength{\tabcolsep}{3pt}
\begin{tabular}{rr} 
\toprule
\textbf{Method} & \textbf{AUC-PR} \\ \midrule 
STL & 0.825 \\
AllTasks & 0.709 \\
ClusMTL & \textbf{0.841} \\
GOMTL & 0.749 \\
Kang & 0.792 \\
Fusion SG-MTL & \textbf{0.837} \\
\bottomrule
\end{tabular}
\caption{TFBS prediction: average AUC-PR, with training data size of 200 examples, test data of $\approx$1800 for number of groups $N$=10.}
\label{tfbs_results}
\end{minipage} \hfill
\begin{minipage}{0.62\textwidth}
\centering
\setlength{\tabcolsep}{3pt}
	\begin{tabular}{cc}
\includegraphics[scale=0.26]{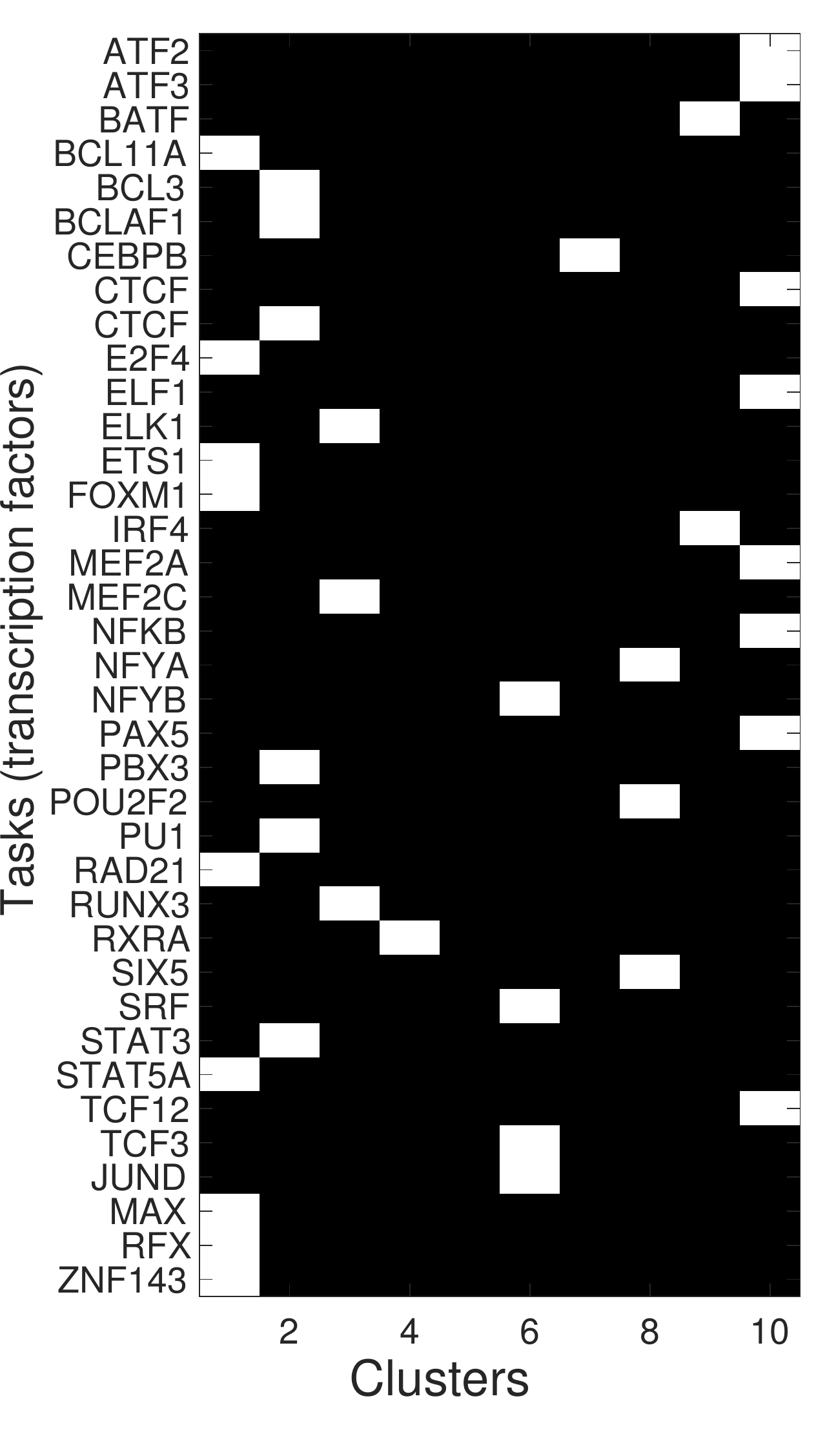} &
\includegraphics[scale=0.28]{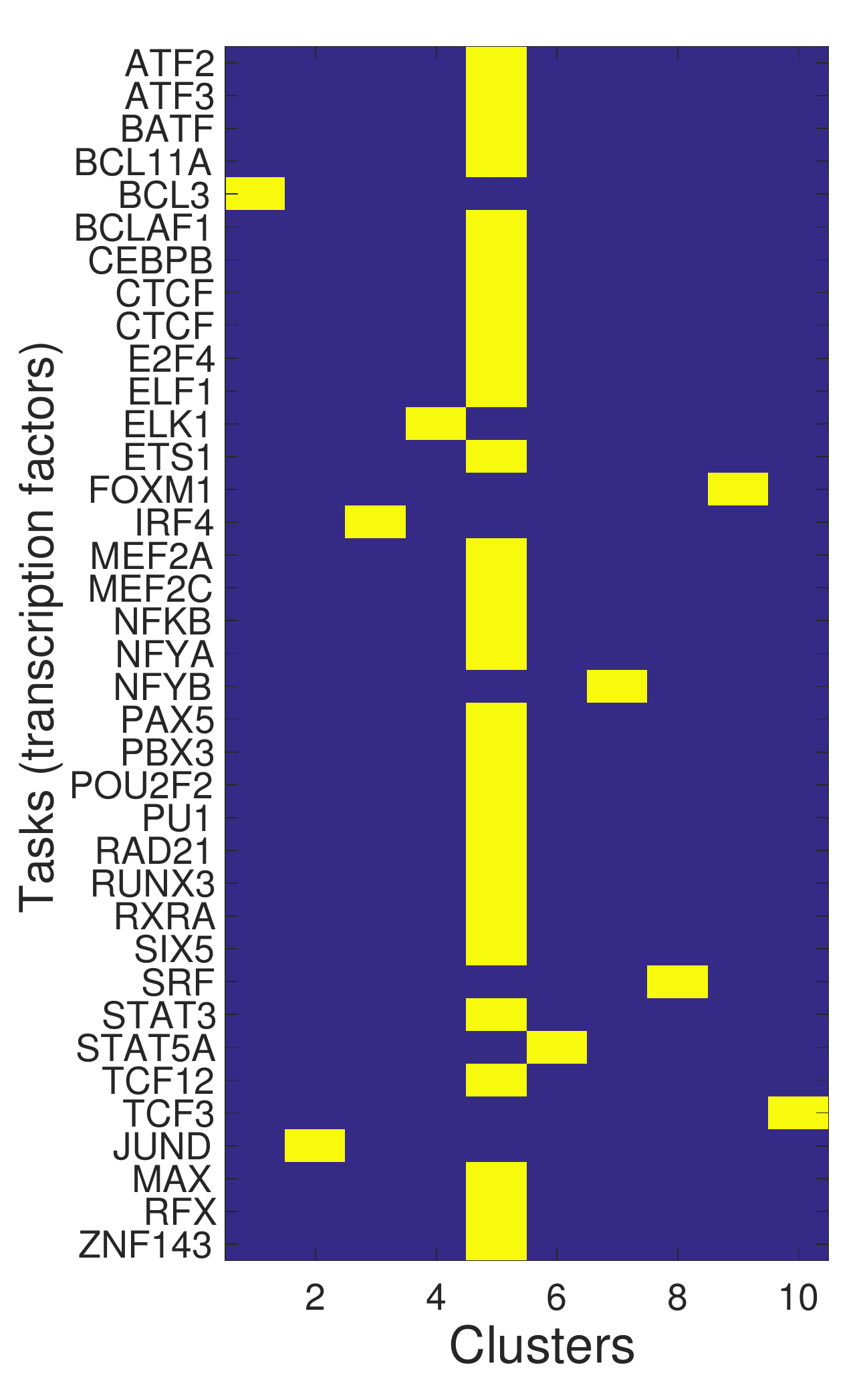} \\
(a) & (b)
   \end{tabular}
\caption{Matrix indicating the groups learned by two methods. The rows show the task names and columns are cluster-ids. A white entry at position $(i,j)$ indicates that task $i$ belongs to cluster $j$ . (a) Groups learned by SG-MTL on the TFBS problem (b) Groups learned by the Clus-MTL baseline show that it tends to put all tasks in the same cluster.}
\label{fig:tfbs_groups}
\end{minipage}
\end{table}

\subsection{Transcription Factor Binding Site Prediction (TFBS)}
This dataset was constructed from processed ChIP-seq data for 37 transcription factors (TFs) downloaded from ENCODE database \citep{encode2012}. Training data is generated in a manner similar to prior literature \citep{seqgl}. Positive examples consist of `peaks' or regions of the DNA with binding events and negatives are regions away from the peaks and called `flanks'. Each of the 37 TFs represents a task. We generate all 8-mer features and select 3000 of these based on their frequency in the data. There are $\approx$2000 examples per task, which we divide into train:test splits using 200 examples (100 positive, 100 negative) as training data and the rest as test data. We report AUC-PR averaged over 5 random train:test splits in Table \ref{tfbs_results}. For our method, we found the number of clusters giving the best AUC-PR to be $N=10$. For the other methods, we tried $N$=5,10,15 and report the best AUC-PR.

Though our method does marginally better (not statistically significant) than the STL baseline, which is a ridge regression model, in many biological applications such as this, it is desirable to have an interpretable model that can produce biological insights. Our MTL approach learns groupings over the TFs which are shown in Fig \ref{fig:tfbs_groups}(a).
Overall, ClusMTL has the best AUC-PR on this dataset however it groups too many tasks into a single cluster (Fig \ref{fig:tfbs_groups}(b)) and forces each group to have at least one task. Note how our method leaves some groups empty (column 5 and 7) as our objective provides a trade-off between adding groups and making groups cohesive.



\section{Conclusion}
We presented a method to learn group structure in multitask learning problems, where the task relationships are unknown. The resulting non-convex problem is optimized by applying the alternating minimization strategy. We evaluate our method through experiments on both synthetic and real-world data. On synthetic data with known group structure, our method outperforms the baselines in recovering them. On real data, we obtain a better performance while learning intuitive groupings. Code is available at: \url{https://github.com/meghana-kshirsagar/treemtl/tree/groups}\\
\noindent Full paper with appendix is available at: \url{https://arxiv.org/abs/1705.04886}

\hspace{-10cm}

\noindent\textbf{Acknowledgements}
We thank Alekh Agarwal for helpful discussions regarding Section \ref{sec:complexity}. E.Y. acknowledges the support of MSIP/NRF (National Research Foundation of Korea) via NRF-2016R1A5A1012966 and MSIP/IITP (Institute for Information \& Communications Technology Promotion of Korea) via ICT R\&D program 2016-0-00563, 2017-0-00537.

\begin{footnotesize}
\bibliography{MTL_GroupLearning}

\begin{thebibliography}{30}
\providecommand{\natexlab}[1]{#1}
\providecommand{\url}[1]{\texttt{#1}}
\expandafter\ifx\csname urlstyle\endcsname\relax
  \providecommand{\doi}[1]{doi: #1}\else
  \providecommand{\doi}{doi: \begingroup \urlstyle{rm}\Url}\fi

\bibitem[Agarwal et~al.(2010)Agarwal, Gerber, and Daume]{agarwal2010}
Agarwal, Arvind, Gerber, Samuel, and Daume, Hal.
\newblock Learning multiple tasks using manifold regularization.
\newblock In \emph{Advances in neural information processing systems}, pp.\
  46--54, 2010.

\bibitem[Argyriou et~al.(2008)Argyriou, Evgeniou, and Pontil]{argyriou2008}
Argyriou, A., Evgeniou, T., and Pontil, M.
\newblock Convex multi-task feature learning.
\newblock \emph{Machine Learning}, 2008.

\bibitem[Bach et~al.(2011)Bach, Jenatton, Mairal, Obozinski, et~al.]{bach2011}
Bach, Francis, Jenatton, Rodolphe, Mairal, Julien, Obozinski, Guillaume, et~al.
\newblock Convex optimization with sparsity-inducing norms.
\newblock \emph{Optimization for Machine Learning}, 5:\penalty0 19--53, 2011.

\bibitem[Baxter(2000)]{baxter2000}
Baxter, Jonathan.
\newblock A model of inductive bias learning.
\newblock \emph{J. Artif. Intell. Res.(JAIR)}, 12:\penalty0 149--198, 2000.

\bibitem[Caruana(1997)]{caruana1997}
Caruana, Rich.
\newblock Multitask learning.
\newblock \emph{Mach. Learn.}, 28\penalty0 (1):\penalty0 41--75, July 1997.
\newblock ISSN 0885-6125.

\bibitem[Chen et~al.(2012)Chen, Liu, and Ye]{chen2012}
Chen, Jianhui, Liu, Ji, and Ye, Jieping.
\newblock Learning incoherent sparse and low-rank patterns from multiple tasks.
\newblock \emph{ACM Transactions on Knowledge Discovery from Data (TKDD)},
  5\penalty0 (4):\penalty0 22, 2012.

\bibitem[Chen \& Ye(2011)Chen and Ye]{Chen2011}
Chen, Yunmei and Ye, Xiaojing.
\newblock Projection onto a simplex.
\newblock \emph{arXiv preprint arXiv:1101.6081}, 2011.

\bibitem[Consortium et~al.(2012)]{encode2012}
Consortium, ENCODE~Project et~al.
\newblock An integrated encyclopedia of dna elements in the human genome.
\newblock \emph{Nature}, 489\penalty0 (7414):\penalty0 57--74, 2012.

\bibitem[Daum{\'e}~III(2009)]{daume2009}
Daum{\'e}~III, Hal.
\newblock Bayesian multitask learning with latent hierarchies.
\newblock In \emph{Proceedings of the Conference on Uncertainty in Artificial
  Intelligence}, pp.\  135--142. AUAI Press, 2009.

\bibitem[Evgeniou \& Pontil(2004)Evgeniou and Pontil]{pontil2004}
Evgeniou, T. and Pontil, M.
\newblock Regularized multi-task learning.
\newblock \emph{ACM SIGKDD}, 2004.

\bibitem[Fei \& Huan(2013)Fei and Huan]{fei2013}
Fei, Hongliang and Huan, Jun.
\newblock Structured feature selection and task relationship inference for
  multi-task learning.
\newblock \emph{Knowledge and information systems}, 35\penalty0 (2):\penalty0
  345--364, 2013.

\bibitem[Gong et~al.(2012)Gong, Ye, and Zhang]{gong2012}
Gong, Pinghua, Ye, Jieping, and Zhang, Changshui.
\newblock Robust multi-task feature learning.
\newblock In \emph{Proceedings of the 18th ACM SIGKDD international conference
  on Knowledge discovery and data mining}, pp.\  895--903. ACM, 2012.

\bibitem[Jacob et~al.(2009)Jacob, Vert, and Bach]{jacob2009}
Jacob, L., Vert, J.P., and Bach, F.R.
\newblock Clustered multi-task learning: A convex formulation.
\newblock In \emph{Advances in neural information processing systems (NIPS)},
  pp.\  745--752, 2009.

\bibitem[Jalali et~al.(2010)Jalali, Sanghavi, Ruan, and Ravikumar]{jalali2010}
Jalali, Ali, Sanghavi, Sujay, Ruan, Chao, and Ravikumar, Pradeep~K.
\newblock A dirty model for multi-task learning.
\newblock \emph{Advances in Neural Information Processing Systems}, pp.\
  964--972, 2010.

\bibitem[Kang et~al.(2011)Kang, Grauman, and Sha]{kang2011}
Kang, Zhuoliang, Grauman, Kristen, and Sha, Fei.
\newblock Learning with whom to share in multi-task feature learning.
\newblock In \emph{International Conference on Machine learning (ICML)}, 2011.

\bibitem[Kim \& Xing(2010)Kim and Xing]{kim2010}
Kim, Seyoung and Xing, Eric~P.
\newblock Tree-guided group lasso for multi-task regression with structured
  sparsity.
\newblock \emph{The Proceedings of the International Conference on Machine
  Learning (ICML)}, 2010.

\bibitem[Kumar \& Daume(2012)Kumar and Daume]{kumar2012}
Kumar, Abhishek and Daume, Hal.
\newblock Learning task grouping and overlap in multi-task learning.
\newblock In \emph{ICML}, 2012.

\bibitem[Liu et~al.(2009)Liu, Ji, and Ye]{liu2009}
Liu, Jun, Ji, Shuiwang, and Ye, Jieping.
\newblock Multi-task feature learning via efficient $l_{2,1}$-norm
  minimization.
\newblock In \emph{Proceedings of the twenty-fifth conference on uncertainty in
  artificial intelligence (UAI)}, pp.\  339--348, 2009.

\bibitem[Ma et~al.(2015)Ma, Sheridan, Liaw, Dahl, and Svetnik]{ma2015}
Ma, Junshui, Sheridan, Robert~P, Liaw, Andy, Dahl, George~E, and Svetnik,
  Vladimir.
\newblock Deep neural nets as a method for quantitative structure--activity
  relationships.
\newblock \emph{Journal of chemical information and modeling}, 55\penalty0
  (2):\penalty0 263--274, 2015.

\bibitem[Maurer(2006)]{maurer2006}
Maurer, Andreas.
\newblock Bounds for linear multi-task learning.
\newblock \emph{The Journal of Machine Learning Research}, 7:\penalty0
  117--139, 2006.

\bibitem[Negahban \& Wainwright(2011)Negahban and Wainwright]{negahban2011}
Negahban, Sahand and Wainwright, Martin~J.
\newblock Estimation of (near) low-rank matrices with noise and
  high-dimensional scaling.
\newblock \emph{The Annals of Statistics}, pp.\  1069--1097, 2011.

\bibitem[Passos et~al.(2012)Passos, Rai, Wainer, and Daume~III]{passos2012}
Passos, Alexandre, Rai, Piyush, Wainer, Jacques, and Daume~III, Hal.
\newblock Flexible modeling of latent task structures in multitask learning.
\newblock \emph{The Proceedings of the International Conference on Machine
  Learning (ICML)}, 2012.

\bibitem[Rao et~al.(2013)Rao, Cox, Nowak, and Rogers]{rao2013sparse}
Rao, Nikhil, Cox, Christopher, Nowak, Rob, and Rogers, Timothy~T.
\newblock Sparse overlapping sets lasso for multitask learning and its
  application to fmri analysis.
\newblock In \emph{Advances in neural information processing systems}, pp.\
  2202--2210, 2013.

\bibitem[Setty \& Leslie(2015)Setty and Leslie]{seqgl}
Setty, Manu and Leslie, Christina~S.
\newblock Seqgl identifies context-dependent binding signals in genome-wide
  regulatory element maps.
\newblock \emph{PLoS Comput Biol}, 11\penalty0 (5):\penalty0 e1004271, 2015.

\bibitem[Tibshirani(1996)]{tibshirani1996}
Tibshirani, Robert.
\newblock Regression shrinkage and selection via the lasso.
\newblock \emph{Journal of the Royal Statistical Society. Series B
  (Methodological)}, pp.\  267--288, 1996.

\bibitem[Widmer et~al.(2010)Widmer, Leiva, Altun, and Ratsch]{widmer}
Widmer, C., Leiva, J., Altun, Y., and Ratsch, G.
\newblock Leveraging sequence classification by taxonomy-based multitask
  learning.
\newblock \emph{RECOMB}, 2010.

\bibitem[Yu et~al.(2005)Yu, Tresp, and Schwaighofer]{yu2005}
Yu, Kai, Tresp, Volker, and Schwaighofer, Anton.
\newblock Learning gaussian processes from multiple tasks.
\newblock In \emph{Proceedings of the 22nd international conference on Machine
  learning}, pp.\  1012--1019. ACM, 2005.

\bibitem[Yuan \& Lin(2006)Yuan and Lin]{yuan2006}
Yuan, Ming and Lin, Yi.
\newblock Model selection and estimation in regression with grouped variables.
\newblock \emph{Journal of the Royal Statistical Society: Series B (Statistical
  Methodology)}, 68\penalty0 (1):\penalty0 49--67, 2006.

\bibitem[Zhang \& Schneider(2010)Zhang and Schneider]{yizhang2010}
Zhang, Yi and Schneider, Jeff~G.
\newblock Learning multiple tasks with a sparse matrix-normal penalty.
\newblock In \emph{Advances in Neural Information Processing Systems}, pp.\
  2550--2558, 2010.

\bibitem[Zhang \& Yeung(2010)Zhang and Yeung]{zhang2010}
Zhang, Yu and Yeung, Dit-Yan.
\newblock A convex formulation for learning task relationships in multi-task
  learning.
\newblock 2010.

\end{thebibliography}
\bibliographystyle{icml2017}
\end{footnotesize}

\newpage
\appendix

\onecolumn

\section{Proof of proposition~\ref{PropSumL1}}
\begin{proof}
	Suppose some tasks $\widehat{\U}_s$ and $\widehat{\U}_t$ in different groups share at least one nonzero patterns. Then, it can be trivially shown that $\sum_{g \in \G} \lambda_g \big\| \W \U_g \big\|_{1,2}$ term will decrease if we combine these two groups because $|a|+|b| > \sqrt{a^2+b^2}$ for any nonzero real number $a$ and $b$. Since $\widehat{\U}$ is fixed, the overall loss will decrease as well.
\end{proof}

\section{Proof of Proposition~\ref{PropSumGN}}
\begin{proof}
	We show the simplest case when $\alpha=2$. Suppose such tasks $\widehat{\U}_s$ and $\widehat{\U}_t$ in the statement. Then, it can be trivially shown that $\sum_{g \in \G} \lambda_g \big\| \W \U_g \big\|_{1,2}$ term will decrease if we split these two tasks in different groups (we can change the group index either task $s$ or $t$ into empty group) because $(|\widehat{w}^{(s)}_i|+|\widehat{w}^{(s)}_j|)^2+(|\widehat{w}^{(t)}_i|+|\widehat{w}^{(t)}_j|)^2 < \Big(\sqrt{(\widehat{w}^{(s)}_i)^2+(\widehat{w}^{(t)}_i)^2}+\sqrt{(\widehat{w}^{(s)}_j)^2+(\widehat{w}^{(t)}_j)^2}\Big)^2$ unless $\widehat{w}^{(s)}_i \widehat{w}^{(t)}_j \neq \widehat{w}^{(s)}_j \widehat{w}^{(t)}_i$. Since $\widehat{\U}$ is fixed, the overall loss will decrease as well.
\end{proof}

\section{Gradient of regularizer terms from \eqref{EqnGrpLearn3}}

For $j=1,2,\hdots,d$, we define a row vector $\W_j \in \reals^{1\times m}$ to denote the $j$-th row of the parameter matrix $\W$: $\W_j = (w^{(1)}_j, w^{(2)}_j, \hdots, w^{(m)}_j)$. Then, with the definition of $\|\cdot\|_{1,2}$, we can rewrite the regularization terms:
\begin{align*}
	R(\U)= & \sum_{g \in \G} \lambda_g \Big( \sum_{j=1}^d   \big \| \W_j \sqrt{\U_g} \big \|_2  \Big)^2 .
\end{align*}
Recalling that that $\sqrt{\U_g}= \diag(\sqrt{u_{g,1}}, \sqrt{u_{g,2}}, \hdots, \allowbreak \sqrt{u_{g,m}})$ where $u_{g,t} \in [0,1]$ and for any fixed task $T_t$, $\sum_{g \in \G} u_{g,t}=1$, the gradient with respect to $u_{g,t}$ can be computed as 
\begin{align*}
	\nabla_{u_{g,t}} R(\U) = \lambda_g  \Big( \sum_{j=1}^d   \big \| \W_j \sqrt{\U_g} \big \|_2  \Big) \Big( \sum_{j=1}^d \frac{ (w_j^{(t)})^2}{\big \| \W_j \sqrt{\U_g} \big \|_2} \Big) .
\end{align*}
The gradient computation of \eqref{EqnGrpLearnFus} with the fusion term can be trivially extended since the squared Frobenius norm is uniformly differentiable; only the additional $2\sum_{t' \neq t} (u_{g,t} - u_{g,t'})$ term needs to be added in $\nabla_{u_{g,t}} R(\U)$ above.

\section{Optimization: gradient of $\L$ and soft-thresholding}
\eqref{EqnCoordW2} can be efficiently solved by proximal gradient descent followed by an element-wise soft thresholding with:
\begin{align}\label{EqnCoordGrad}
	& \nabla \widebar{\L} := \nabla_{w_j^{(t)}} \L(w_j^{(t)}) + 2 \sum_{g \in \G} \lambda_g  w_j^{(t)} + \\
	& 2 \sum_{g \in \G^+} \lambda_g \bigg(\sum_{j' \neq j} \sqrt{\sum_{t'=1}^m  u_{g,t'} (w_{j'}^{(t')})^2 } \bigg) \frac{u_{g,t} w_j^{(t)}}{\sqrt{\sum_{t'=1}^m  u_{g,t'} (w_j^{(t')})^2}} , \nonumber
\end{align}
followed by an element-wise soft thresholding $\mathcal{S}_{\nu} (a) := \text{sign} (a) \max(|a| - \nu,0)$. The amount of soft-thresholding is determined by the learning rate $\eta$ and the constant factor of $| w_j^{(t)}|$ which we call $\bar{\lambda}$:
\begin{align}\label{EqnCoordLambda}
	\bar{\lambda} : = 2 \sum_{g \in \G^0} \lambda_g \bigg(\sum_{j' \neq j} \sqrt{\sum_{t'=1}^m  u_{g,t'} (w_{j'}^{(t')})^2 } \bigg) \sqrt{u_{g,t}}.
\end{align}

\section{Interpretation of clusters learned}

\subsection{QSAR task clusters}
\begin{table}
\caption{15 tasks corresponding to the 15 drug targets (target name and description is shown) and their groupings. The color of the cell indicates the assigned cluster.}
\label{tasks_drugs}
{\footnotesize
\begin{tabular}{p{3.7cm}|p{3.7cm}} 
\toprule
3A4: CYP P450 3A4 inhibition & \cellcolor{blue!25}CB1: binding to cannabinoid receptor 1\\\hline
\cellcolor{blue!25}DPP4: inhibition of dipeptidyl peptidase 4& \cellcolor{green!25}HIVINT: inhibition of HIV integrase in a cell based assay \\ \hline
\cellcolor{cyan!25}HIVPROT: inhibition of HIV protease & \cellcolor{blue!25}LOGD: lipophilicity measured by HPLC method\\ \hline
\cellcolor{blue!25}OX1: inhibition of orexin 1 receptor & \cellcolor{green!25}METAB: percent remaining after 30 min microsomal incubation  \\ \hline
\cellcolor{blue!25}OX2: inhibition of orexin 2 receptor & \cellcolor{blue!25}NK1: inhibition of neurokinin1 receptor binding \\ \hline
\cellcolor{red!25}PGP: transport by p-glycoprotein& \cellcolor{red!25}PPB: human plasma protein binding \\ \hline
\cellcolor{yellow!25}RATF: rat bioavailability & \cellcolor{cyan!25}TDI: time dependent 3A4 inhibitions\\ \hline
\cellcolor{cyan!25}THROMBIN : human thrombin inhibition & \\
\bottomrule
\end{tabular}
}
\end{table}

For the results in Table \ref{qsar_results}, the number of task clusters $N$ is set to 5 (for the baseline methods we tried $N$=2,5,7). We found that Clus-MTL tends to put all tasks in the same cluster for any value of $N$. Our method has the lowest average MSE.
In Figure \ref{fig:learning_curve}, for our method we show how MSE changes with the number of groups $N$ (along x-axis) and over different sizes of the training/test split. The dip in MSE for $n=300$ (green box curve) between $N=5$ to $N=7$ suggests there are around 5-7 groups.  

In Fig \ref{tasks_drugs} we show the grouping of the 15 targets. Please note that since we do not use all features available for each task (we only selected the common 3000 features among all 15 tasks), the clustering we get is likely to be approximate. We show the clustering corresponding to a run with Avg MSE of 0.6082 (which corresponds to the dip in the MSE curve from Fig \ref{fig:learning_curve}). We find that CB1, OX1, OX2, NK1 which are all receptors corresponding to neural pathways, are put in the same cluster (darker blue). OX1, OX2 are targets for sleep disorders, cocaine addiction. NK1 is a G protein coupled receptor (GPCR) found in the central nervous system and peripheral nervous system and has been targetted for controlling nausea and vomiting. CB1 is also a GPCR involved in a variety of physiological processes including appetite, pain-sensation, mood, and memory. In addition to these, DPP4 is also part of the same cluster; this protein is expressed on the surface of most cell types and is associated with immune regulation, signal transduction and apoptosis. Inhibitors of this protein have been used to control blood glucose levels. LOGD is also part of this cluster: lipophilicity is a general physicochemical property that determines binding ability of drugs. We are not sure what this task represents. 

PGP and PPB are related to plasma and are in the same group. Thrombin, a protease and a blood-coagulant protein, HIV protease and 3A4 (which is involved in toxin removal) are in the same group. Further analysis of the features might reveal the commonality between mechanisms used to target these proteins.

\end{document}